\documentclass[10pt,twocolumn,letterpaper]{article}

\usepackage{cvpr}              % To produce the CAMERA-READY version

\usepackage[accsupp]{axessibility}  % Improves PDF readability for those with disabilities.

\usepackage{titletoc}
\usepackage{float}
\usepackage{placeins}

\usepackage{xcolor}
\usepackage[T1]{fontenc}
\definecolor{my_green}{RGB}{51,102,0}
\definecolor{my_red}{RGB}{204, 0, 0}
\definecolor{paired-light-blue}{RGB}{198, 219, 239}
\definecolor{paired-dark-blue}{RGB}{49, 130, 188}
\definecolor{paired-light-orange}{RGB}{251, 208, 162}
\definecolor{paired-dark-orange}{RGB}{230, 85, 12}
\definecolor{paired-light-green}{RGB}{199, 233, 193}
\definecolor{paired-dark-green}{RGB}{49, 163, 83}
\definecolor{paired-light-purple}{RGB}{218, 218, 235}
\definecolor{paired-dark-purple}{RGB}{117, 107, 176}
\definecolor{paired-light-gray}{RGB}{217, 217, 217}
\definecolor{paired-dark-gray}{RGB}{99, 99, 99}
\definecolor{paired-light-pink}{RGB}{222, 158, 214}
\definecolor{paired-dark-pink}{RGB}{123, 65, 115}
\definecolor{paired-light-red}{RGB}{231, 150, 156}
\definecolor{paired-dark-red}{RGB}{131, 60, 56}
\definecolor{paired-light-yellow}{RGB}{231, 204, 149}
\definecolor{paired-dark-yellow}{RGB}{141, 109, 49}  
\definecolor{myblue}{RGB}{218,232,252}
\definecolor{mygray}{RGB}{220,220,220}
\definecolor{mypink}{RGB}{251,49,153}

\newcommand{\myparagraph}[1]{\textbf{#1}\hspace{1.8ex}}

\usepackage{colortbl}
\usepackage{enumitem}

\definecolor{cvprblue}{rgb}{0.21,0.49,0.74}
\usepackage[colorlinks,
            linkcolor=black,
            anchorcolor=blue,
            citecolor=purple,
            ]{hyperref}

\def\paperID{2503} % *** Enter the Paper ID here
\def\confName{CVPR}
\def\confYear{2025}

\title{Identity-Preserving Text-to-Video Generation by Frequency Decomposition}

\author{%
  Shenghai Yuan\textsuperscript{1,4},
  Jinfa Huang\textsuperscript{3},
  Xianyi He\textsuperscript{1,4},
  Yunyang Ge\textsuperscript{1,4},
  \\
  Yujun Shi\textsuperscript{5},
  Liuhan Chen\textsuperscript{1,4},
  Jiebo Luo\textsuperscript{3},
  Li Yuan\textsuperscript{1,2,†}
  \\
  \\
  \textsuperscript{1} Peking University,
  \textsuperscript{2} Peng Cheng Laboratory,
  \textsuperscript{3} University of Rochester,
  \\
  \textsuperscript{4} Rabbitpre Intelligence,
  \textsuperscript{5} National University of Singapore
  \\
  \\
  {
    \tt 
    \small
    \{yuanshenghai,HeXianyi,yunyang,chenliuhan\}@stu.pku.edu.cn,
    shi.yujun@u.nus.edu,
  } 
  \\
  {
    \tt 
    \small
    yuanli-ece@pku.edu.cn,
    \{jhuang90@ur,jluo@cs\}.rochester.edu
  } 
  \\
  \\
  \textbf{\url{https://github.com/PKU-YuanGroup/ConsisID}}
  \\
}  

\clearpage
\begin{document}

\twocolumn[{%
\maketitle
\begin{figure}[H]
\vspace{-10mm}
        \hsize=\textwidth
        \centering
        \includegraphics[width=1.96\linewidth]{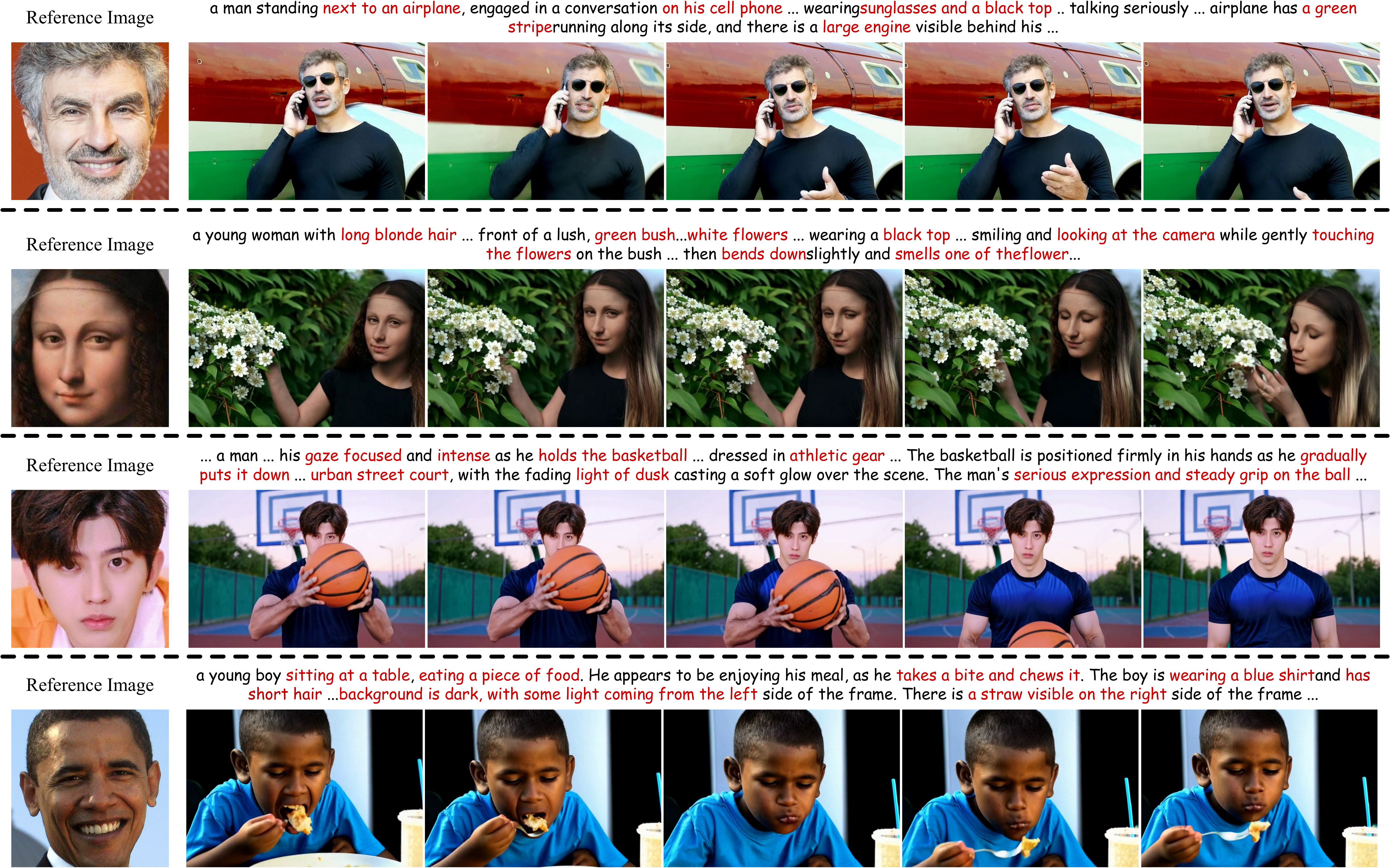}
        \caption{\textbf{Examples of identity-preserving video generation (IPT2V) by our ConsisID.} Given a reference image, our method can generate realistic and personalized human-centered videos while preserving identity. \textcolor{Red}{\textbf{Red}} indicates that attributes in long instructions.}
        \label{figure_main_results_1}
        % \vspace{8pt}
\end{figure}
}]

\begin{abstract}

Identity-preserving text-to-video (IPT2V) generation aims to create high-fidelity videos with consistent human identity.  It is an important task in video generation but remains an open problem for generative models. This paper pushes the technical frontier of IPT2V in two directions that have not been resolved in the literature: (1) A tuning-free pipeline without tedious case-by-case finetuning, and (2) A frequency-aware heuristic identity-preserving Diffusion Transformer (DiT)-based control scheme.
To achieve these goals, we propose \textbf{ConsisID}, a tuning-free DiT-based controllable IPT2V model to keep human-\textbf{id}entity \textbf{consis}tent in the generated video. Inspired by prior findings in frequency analysis of vision/diffusion transformers, it employs identity-control signals base on frequency domain, since facial features can be decomposed into low-frequency global features (\eg, profile, proportions) and high-frequency intrinsic features (\eg, identity markers that remain unaffected by pose changes).
Extensive experiments demonstrate that our frequency-aware heuristic scheme provides an optimal control solution for DiT-based models, making strides toward more effective IPT2V. 

\begin{figure*}[ht]
  \centering
  \includegraphics[width=0.9\linewidth]{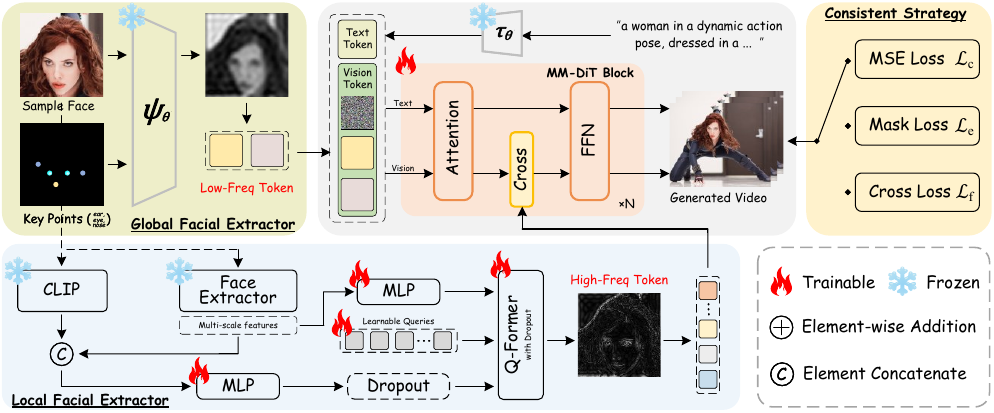}
  \caption{\textbf{Overview of the proposed ConsisID.} Based on \textit{Findings of DiT}, low-frequency facial information is embedded into the shallow layers, while high-frequency information is incorporated into the vision tokens within the attention blocks. The ID-preserving Recipe is applied to ease training and improve generalization. The \textit{cross face}, \textit{DropToken}, and \textit{Dropout} are executed based on probability.}
  \label{fig: model_pipeline}
\end{figure*}

\end{abstract}
    
\section{Introduction}\label{sec: introduction}

Large-scale pre-trained video diffusion models \cite{opensoraplan, opensora, allegro, cogvideox} have facilitated a variety of downstream applications \cite{magictime, controlnet, InstaDrag, evagaussians, cycle3d, ViewCrafter}, particularly in identity-preserving text-to-video (IPT2V) \cite{dreamvideo, customvideo, motionbooth, Still-moving, magic-me}. However, existing methods face significant challenges, particularly the high overhead associated with the need for case-by-case finetuning, which diminishes their applicability. Within the open-source community, only the ID-Animator \cite{ID-Animator} can implement tuning-free IPT2V, but it can only generate videos similar to talking head \cite{TalkingHead-1KH} and has poor id preservation. 

Additionally, the above efforts are predominantly based on U-Net and cannot be adapted to the emerging DiT-based video model \cite{opensoraplan, opensora, cogvideox, allegro, easyanimate}. This challenge may stem from the inherent limitations of DiT compared to U-Net, including greater difficulty in training convergence and weakness in perceiving facial details. From some prior findings in frequency analysis of vision/diffusion transformers \cite{rethinking_DiT, ViT, HAT, U-ViT, U-DiTs, Freeu, yu2024promptfix}, we can know that the reason is: \textbf{Finding 1:} \textit{Shallow (\eg, low-level, low-frequency) features are essential for pixel-level prediction tasks in diffusion models, as they ease model training.} U-Net facilitates model convergence by aggregating shallow features to the decoder via long skip connections, a mechanism that DiT does not incorporate; \textbf{Finding 2:} \textit{Transformers have limited perception of high-frequency information, which is important for preserving facial features.} The encoder-decoder architecture of U-Net naturally possesses multi-scale features (\eg, richness in high-frequency), while DiT lacks a comparable structure. To develop a DiT-based control model, these must be addressed first. Please see Appendix for more details.

For ID-preserving video generation, the challenges stem from the requirement for each frame to incorporate both high-frequency (\eg, age- and make-up-independent identity markers) and low-frequency information (\eg, facial shape) derived from the reference image, which can just be used to make up for the DiT defects mentioned above. Therefore, we propose \textbf{ConsisID}, to keep the \textbf{identity} \textbf{consis}tency in video generation by frequency decomposition, based on the previously \textbf{Findings of DiT} in frequency analysis. Thanks to the large-scale pre-trained DiT, we can use its powerful capabilities to achieve tuning-free effects. ConsisID decouples identity features into high- and low-frequency signals, which are injected into specific locations within the DiT, facilitating efficient IPT2V generation. Specifically, in line with \textit{Finding 1}, we first convert the reference image and the facial key points to the low-frequency signal, then concatenate them with input noise latent to ease the training. Following \textit{Finding 2}, we utilize a dual-tower feature extractor to capture high-frequency facial information, which is integrated with vision tokens within the transformer block, thereby enhancing the DiT's high-frequency perception capabilities. Finally, to transform the pre-trained model into an IPT2V model and improve its generalization, we further introduce a hierarchical training strategy. 

Our contributions can be summarized as follows:

\begin{itemize}
    \item We introduce \textbf{ConsisID}, a tuning-free identity-preserving DiT-based IPT2V model, which preserves the identity of the main subject of the video using control signals from frequency decomposition. 
    \item We propose a hierarchical training strategy, including coarse-to-fine training, dynamic mask loss, and dynamic cross-face loss, which work together to facilitate training and enhance generalization effectively.
    \item Extensive experiments demonstrate that our \textbf{ConsisID} can generate high-quality, editable, consistent identity-preserving videos, benefiting from our frequency-aware identity-preserving T2V DiT-based control scheme.
\end{itemize}

\section{Related Work}\label{sec: related work}
\myparagraph{Tuning-based Identity-preserving T2V Models.}
Diffusion models are widely recognized for their strong generative capabilities \cite{cfree-guidance, DALLE, Followyourpose, Follow-your-click, Follow-your-emoji, Magicstick, magictime, chronomagic-bench}, significantly advancing the development of identity-preserving generative models \cite{Elite, Photoverse, controlnet, t2iadapter}. Initially, the researchers used tuning-based methods to generate content that matched the input ID. This process requires finetuning pretrained model for each new person during inference. For example, DreamBooth \cite{Dreambooth} introduced a novel loss function to fine-tune the entire network, embedding identity information while preserving the original generative capabilities. LoRA \cite{Lora}, similar to DreamBooth \cite{Dreambooth}, requires training only a small subset of network parameters. In contrast, Textual Inversion \cite{textual_inversion} freezes the pretrained network and embeds identity information into a trainable word embedding. Subsequent tuning-based methods, including both image and video models based on U-Net or DiT architectures \cite{Hyperdreambooth, multi-concept, dreamvideo, customvideo, motionbooth, Still-moving, xiong2024autoregressive, yang2024parameter, zheng2024videogen}, generally follow three main approaches. While these models demonstrate substantial effectiveness, the requirement to fine-tune for each new identity restricts their practical applicability.

\noindent\myparagraph{Tuning-free Identity-preserving T2V Models.}
To address the issue of high resource consumption, several tuning-free diffusion models have recently emerged in the field of image generation \cite{Ip-adapter, instantid, UniPortrait, pulid, photomaker}. These models do not require finetuning parameters for newly introduced IDs during inference. For instance, IP-Adapter \cite{Ip-adapter} utilizes the CLIP \cite{CLIP} features of the identity image through cross-attention to guide the pretrained model in generating identity-preserving images. InstantID \cite{instantid} extends this approach by replacing CLIP \cite{CLIP} features with Arcface \cite{arcface} features and integrating a pose network to adjust facial proportions. Unlike these initial methods, which introduce control signals via visual tokens, PhotoMaker \cite{photomaker} and Imagine Yourself \cite{Imagine_yourself} leverage text tokens. Specifically, PhotoMaker \cite{photomaker} concatenates identity features obtained from the CLIP encoder \cite{CLIP} to the text embedding, while Imagine Yourself \cite{Imagine_yourself} uses element-wise addition for feature fusion. In the domain of video generation, only MovieGen \cite{movie_gen} and ID-Animator \cite{ID-Animator} currently support ID-preserving text-to-video (IPT2V) generation. MovieGen is closed-source, whereas ID-Animator is open-source but uses a methodology similar to image models, leading to lower-quality identity preservation in the generated videos. We select the emerging DiT architecture \cite{Latte, opensoraplan, opensora, cogvideox} and optimize it for IPT2V, drawing on conclusions from prior frequency analyses \cite{rethinking_DiT, ViT, HAT, U-ViT, U-DiTs, Freeu}. This enables high-quality, editable, and consistent ID-preserving video generation.

% While most existing works rely on the U-Net architecture \cite{nuwa, animatediff, Magicvideo}, we adopt the emerging DiT architecture \cite{Latte, opensoraplan, opensora, cogvideox}. We provide a novel analysis of this architecture from a frequency domain perspective \cite{rethinking_DiT, ViT, HAT, U-ViT, U-DiTs, Freeu}, decoupling ID features into high- and low-frequency control signal, which are then injected into specific positions in DiT \cite{DiT, SiT}. This enables efficient and interpretable DiT-based ID-preserving video generation.
\section{Methodology}\label{sec: method}

\subsection{Preliminaries}
\noindent\myparagraph{Diffusion Model.}
Text-to-video generation models usually utilize the diffusion paradigm, which gradually transforms noise $\epsilon$ into a video $x_0$. Originally, denoising was conducted directly within the pixel space \cite{DDIM, DDPM, DPM}; however, due to significant computational overheads, recent methods predominantly employ latent space \cite{LDM, Videopoet, magictime, animatediff}. The optimization process is defined:
\begin{equation}
\label{eq: original_loss_function}
\mathcal{L}_{a}=\mathbb{E}_{x_0, t, y, \epsilon}\left[\left\|\epsilon-\epsilon_\theta\left(x_0, t, \tau_\theta(y)\right)\right\|_2^2\right],
\end{equation}
where $y$ is text condition, $\epsilon$ is sampled from a standard normal distribution (\eg, $\epsilon \sim \mathcal{N}(0, 1)$), and $\tau_\theta(\cdot)$ is the text encoder. By replacing $x_0$ with $\mathcal{E}\left(x_0\right)$, the latent diffusion is derived, which is used by ConsisID.

\noindent\myparagraph{Diffusion Transformer.}
The DiT-based video generation model shows significant potential in simulating the physical world \cite{SORA, cogvideox, allegro}. Despite being a novel architecture, research on controllable generation has been limited, and current methods \cite{tora, pulid, Camera_DiT, movie_gen} largely resemble U-Net based approaches \cite{controlnet, t2iadapter, textual_inversion}. However, no study has yet examined why this approach works with DiT. Drawing from prior analyses of Diffusion and Transformer from a frequency domain perspective \cite{rethinking_DiT, ViT, HAT, U-ViT, U-DiTs, Freeu}, we conclude that: \textbf{(1)} Low-frequency (\eg, shallow-layer) features are essential for pixel-level prediction tasks in diffusion models, which helps facilitate model training; \textbf{(2)} Transformers have limited perception for high-frequency information, which is important for controllable generation. Based on these, we decouple ID features into high- and low-frequency parts and inject them into specific locations, achieving effective identity-preserving text-to-video generation.

\subsection{ConsisID: Keep Your Identity Consistent}
The overview is illustrated in Figure \ref{fig: model_pipeline}. Given a reference image, the global facial extractor and local facial extractor inject both high- and low-frequency facial information into model, which then generates identity-preserving videos with the assistance of the consistency training strategy.

\subsubsection{Low-frequency View: Global Facial Extractor}
In light of \textit{Finding 1}, enhancing low-level (\eg, shallow, low-frequency) features accelerates model convergence. To easily adapt a pre-trained model for the IPT2V task, the most direct approach is concatenating the reference face with the noise input latent \cite{svd}. However, the reference face contains both high-frequency details (\eg, eye and lip textures) and low-frequency information (\eg, facial proportions and contours). From \textit{Finding 2}, prematurely injecting high-frequency information into the Transformer is inefficient and may hinder the model's processing of low-frequency information, as the Transformer focuses primarily on low-frequency features. In addition, feeding the reference face directly into the model could introduce irrelevant noise such as lighting and shadows. To mitigate this, we extract facial key points, convert them to an RGB image, and then concatenate it with the reference image, as shown in Figure \ref{fig: model_pipeline}. This strategy focuses the model's attention on the low-frequency signals in the face, while minimizing the impact of extraneous features. We found that when this component is discarded, the model is difficult to convergen. The objective function is changed to:
{\small
\begin{equation}
\label{eq: global_loss_function}
\mathcal{L}_{b}=\mathbb{E}_{x_0, t, y, f, \epsilon}\left[\left\|\epsilon-\epsilon_\theta\left(x_0, t, \tau_\theta(y), \psi_\theta(f)\right)\right\|_2^2\right],
\end{equation}
}where $\psi_\theta(\cdot)$ is a variational autoencoder, $f$ represents the reference image, we ignore key points here for simplicity.

\subsubsection{High-frequency View: Local Facial Extractor}
In light of \textit{Finding 2}, we recognize that Transformers have limited sensitivity to high-frequency information. It can be concluded that relying solely on global facial extractor is insufficient for IPT2V generation, as global facial features lack of high-frequency information. So we use a face recognition backbone \cite{arcface} to extract high-frequency features, as these are invariant to non-ID attributes (\eg, expression, posture, and shape). We refer to these features as intrinsic identity features (\eg, high-frequency), since age and makeup do not alter an individual’s core identity. Following \cite{UniPortrait}, we utilize the penultimate layer of the backbone, rather than its output, as it retains more spatial information pertinent to identity. However, our experiments reveal that while the face recognition backbone improves identity consistency, it lacks the semantic features required for editing. This task demands not only maintaining identity consistency but also incorporating the ability to edit, such as generating videos of faces with the same identity but varying age and makeup. Previous research \cite{Imagine_yourself, ID-Animator} relies solely on the CLIP encoder \cite{CLIP} to enable editing capabilities. However, since CLIP is not specifically trained on face datasets, the features it extracts include irrelevant non-face information, which can compromise identity fidelity \cite{photomaker, instantid, Ip-adapter}.

To address these challenges, we first use a facial recognition backbone to extract features that strongly represent intrinsic identity, and a CLIP image encoder to capture semantically rich features. We then employ the Q-Former \cite{blip1, blip2, blip3} to fuse these features, resulting in intrinsic identity representations enriched with high-frequency semantic information. To mitigate the impact of irrelevant features from CLIP, dropout \cite{drop_token, dropout} is applied post-processing. Additionally, we follow \cite{UniPortrait} to concatenate the shallow, multi-scale features from the facial recognition backbone (after interpolation) with the CLIP features. This approach ensures that the model captures essential intrinsic identity features while filtering out extraneous noise unrelated to identity. Finally, we apply cross-attention to facilitate interaction between this feature set and the visual tokens produced by each attention block of the pre-trained model, thereby enhancing the high-frequency information in the DiT:
\begin{equation}
Z_i^{\prime} = Z_i + \text{Attention}(Q_i^v, K_i^f, V_i^f),
\end{equation}
where $i$ represents the layer number of the attention block, $Q^{v} = Z_i W^{q}_i$, $K^{f} = F W^{k}_i$, and $V^{f} = F W^{v}_i$, where $Z_i$ is the visual token, $F$ represents the intrinsic identity features, and $W_q$, $W_k$, and $W_v$ are trainable parameters. The objective function is changed to:
{\small
\begin{equation}
\label{eq: local_loss_function}
\mathcal{L}_{c}=\mathbb{E}_{x_0, t, y, f, \epsilon}\left[\left\|\epsilon-\epsilon_\theta\left(x_0, t, \tau_\theta(y), \psi_\theta(f), \varphi_\theta(f)\right)\right\|_2^2\right],
\end{equation}
}where $\varphi_\theta(\cdot)$ is the local facial extractor.

\subsubsection{Consistency Training Strategy}\label{sec: id-oriened training recipe}
During training, we randomly select a frame from the training frames and apply the Crop \& Align \cite{arcface} to extract the facial region as reference images, which is subsequently used as an identity-control signal, alongside the text as control. 

\noindent\myparagraph{Coarse-to-Fine Training.}
Compared to Identity-preserving image generation, video generation requires maintaining consistency in both spatial and temporal dimensions, ensuring that high and low-frequency facial information matches the reference image. To mitigate the complexity of training, we propose a hierarchical strategy where the model learns information globally before refining it locally. In the coarse-grained phase (\eg, corresponding \textit{Finding 1}), we employ the global facial extractor, enabling the model to prioritize low-frequency features, such as facial contours and proportions, thereby ensuring rapid acquisition of identity information from the reference image and consistency across the video sequence. In the fine-grained phase (\eg corresponding to \textit{\textit{Finding 2}}), the local facial extractor shifts the model's focus to high-frequency details, such as the texture details of eyes and lips (\eg, intrinsic identification), improving the fidelity of facial expressions and the overall similarity of the generated face.

\noindent\myparagraph{Dynamic Mask Loss.}
The objective of our task is to ensure that the identity of the person in the generated video remains consistent with the input reference image.  However, Equation \ref{eq: original_loss_function} considers the entire scene, encompassing both high- and low-frequency identity information as well as redundant background content, which introduces noise that interferes with model training. To address this, we propose to focus the model's attention on face regions. Specifically, we first extract the facial mask from the video, apply trilinear interpolation to map it to the latent space, and finally use this mask to constrain the computation of $\mathcal{L}_{c}$:
\begin{equation} 
\label{eq: mask_1_loss_function_mask} 
\mathcal{L}_{d}=M \odot \mathcal{L}_{c},
\end{equation}
where $M$ represents a mask with the same shape as $\epsilon$. However, if Equation \ref{eq: mask_1_loss_function_mask} is used as the supervisory signal for all training data, the model may fail to generate a natural background during inference. To mitigate this issue, we apply Equation \ref{eq: mask_1_loss_function_mask} with a probability $p$ of $\alpha$, resulting in:
\begin{equation}
\label{eq: dynamic_mask_loss}
\mathcal{L}_{e} = 
\begin{cases}
\mathcal{L}_{d}, & \text{if } p > \alpha \\
\mathcal{L}_{c}, & \text{if } p \leq \alpha
\end{cases}
\end{equation}
% where the $\alpha$ is set to 0.5 to achieve the optimal balance.

\begin{figure}[t]
    \centering
    \includegraphics[width=0.85\linewidth]{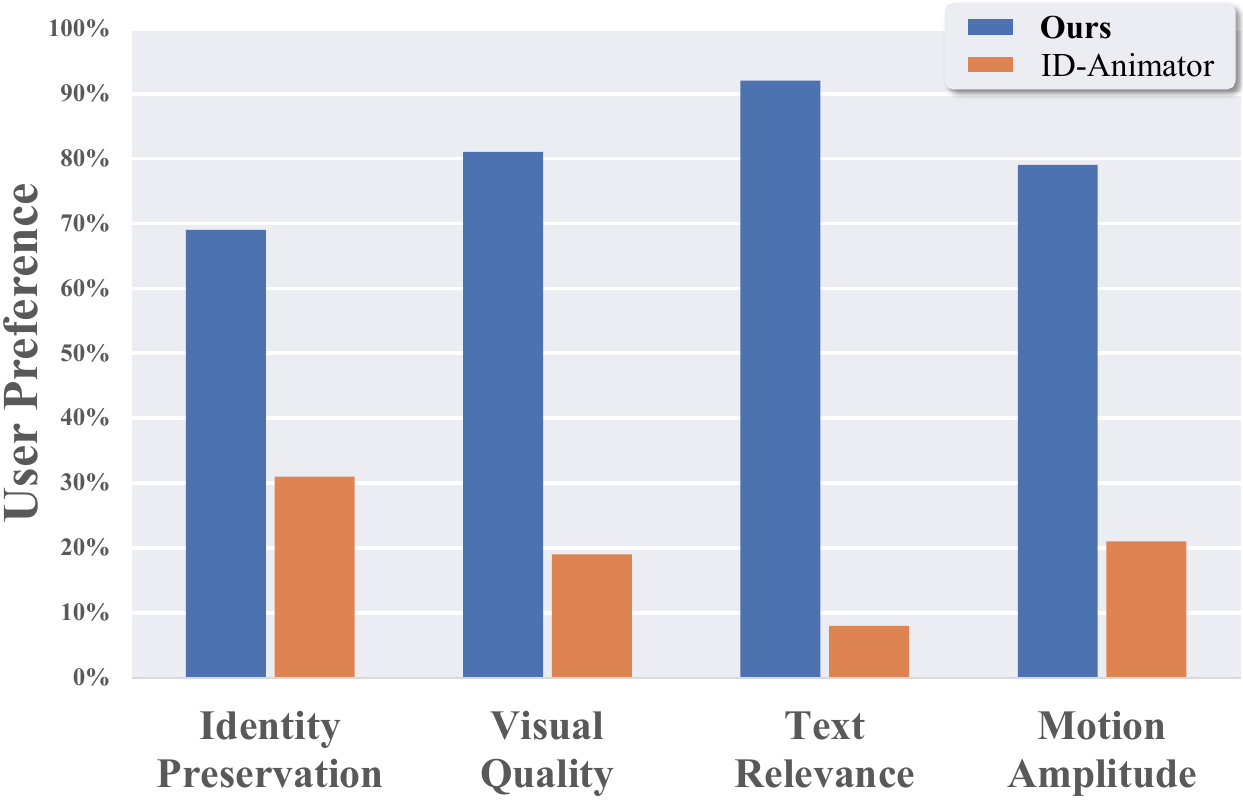}
    \caption{\textbf{User Study between ConsisID and state-of-the-art methods.} ConsisID is preferred by voters in all dimensions.}
    \label{fig: human_evaluation}
    % \vspace{-5pt}
\end{figure}

\noindent\myparagraph{Dynamic Cross-Face Loss.}
After training with Equation \ref{eq: dynamic_mask_loss}, we observed that the model struggled to generate satisfactory results for persons not present in the data domain during inference. This issue arises because the model, trained exclusively on faces from the training frames, tends to overfit by adopting a "copy-paste" shortcut—essentially replicating the reference image without alteration. To improve the model’s generalization capability, we introduce slight Gaussian noise $\zeta$ to the reference images and use cross-face (\eg, reference images are sourced from video frames outside the training frames) as inputs with probability $\beta$:
\begin{equation}
\mathcal{L}_f = 
\begin{cases} 
\mathcal{L}_e \quad \text{where} \quad x_0 \cdot \zeta, & \text{if } p > \beta \\
\mathcal{L}_e \quad \text{where} \quad x_c \cdot \zeta, & \text{if } p \leq \beta
\end{cases}
\end{equation}
% \begin{equation}
% \mathcal{L}_f = \mathcal{L}_e \quad \text{where} \quad x_0 = 
% \begin{cases} 
% x_0 \cdot \zeta, & \text{if } p > \beta \\
% x_c \cdot \zeta, & \text{if } p \leq \beta
% \end{cases}
% \end{equation}
where $x_0$ is the reference image extracted from the training frames, and $x_c$ is extracted from outside the training frames.
% the $\beta$ is set to 0.5 to achieve the optimal balance.

\begin{table}[t]
    \centering
    \resizebox{\columnwidth}{!}
    {
        \begin{tabular}{c|cccc}
            \toprule
             & FaceSim-Arc $\uparrow$ & FaceSim-Cur $\uparrow$ & CLIPScore $\uparrow$ & FID $\downarrow$  \\
            \midrule
            ID-Animator \cite{ID-Animator} & 0.32 & 0.33 & 24.97 & \textbf{117.46} \\
         \rowcolor{myblue}   \textbf{ConsisID} & \textbf{0.58} & \textbf{0.60} & \textbf{27.93} & 151.82 \\
            \bottomrule
        \end{tabular}
    }
    \caption{\textbf{Quantitative comparison with state-of-the-art methods.} ConsisID achieve well-aligned results across most metrics. "$\downarrow$" denotes lower is better. "$\uparrow$" denotes higher is better.}
    \label{tab: quantitative analysis}
    \vspace{-3.8pt}
\end{table}

\begin{figure*}[!t]
    \centering
    \includegraphics[width=0.82\linewidth]{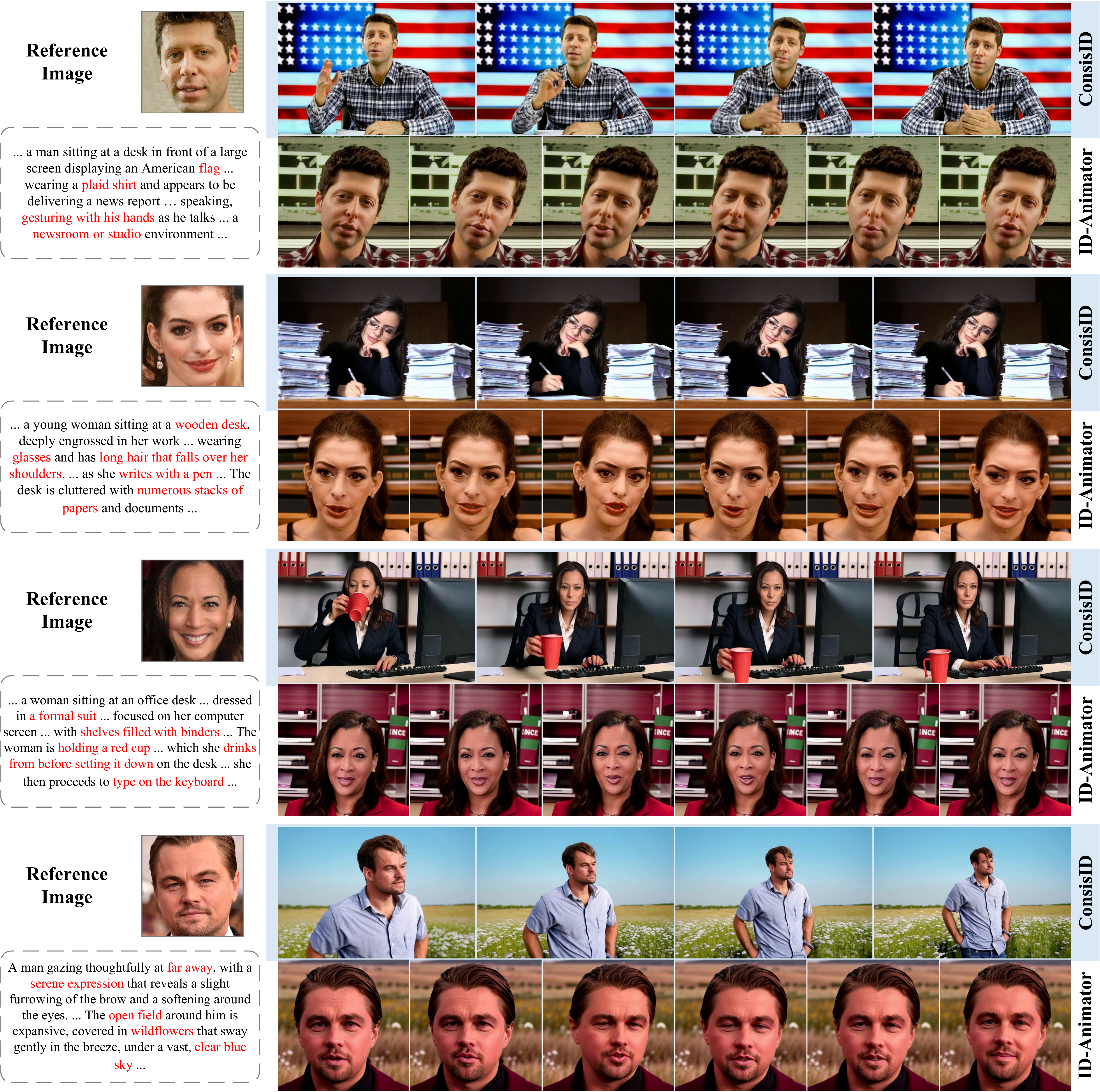}
    \caption{\textbf{Qualitative analysis between ConsisID and ID-Animator \cite{ID-Animator}.} ID-Animator can only generate videos of the face region, and the identity Preservation is poor (\eg, shape, texture). Additionally, it cannot generate specified content according to the text prompt (\eg, action, decoration, background). ConsisID achieves advantages in identity preservation, visual quality, motion amplitude, and text relevance. Moreover, our ConsistID can generate more frames rather than ID-Animator (49 480$\times$720p frames v.s. 16 512$\times$512p frames).}
    \label{fig: qualitative analysis}
    \vspace{-5pt}
\end{figure*}

\section{Experiments}\label{sec: experiment}

\subsection{Setup}\label{sec: setup}
\noindent\myparagraph{Implementation details.}
% ConsisID is compatible with various DiT generative architectures \cite{opensora, opensoraplan, allegro}; 
ConsisID selects DiT-based generation architectures CogVideoX-5B \cite{cogvideox} as our baseline for validation. We use an in-house human-centric dataset for training, which differs from previous datasets \cite{TalkingHead-1KH, celebv-text, voxceleb} that focus only on the face. In the training phase, we set the resolution to 480$\times$720 and extracted 49 consecutive frames at a stride of 3 from each video as training data. We set the batch size to 80, the learning rate to $3\times10^{-6}$, and the total number of training steps to 1.8k. The randomly discarded text rate is set to $0.1$, with AdamW serving as an optimizer and \textit{cosine\_with\_restarts} as a learning rate scheduler. The training strategy is the same as Section~\ref{sec: id-oriened training recipe}. We set $\alpha$ and $\beta$ in the dynamic cross-face loss ($\mathcal{L}_{e}$) and dynamic mask loss ($\mathcal{L}_{f}$) to $0.5$, respectively. In the inference phase, we employ DPM \cite{DPM} with a sampling step of $50$, and a classifier free guidance ratio of $6.0$. For more detials and results, please refer to Appendix. 
%, and maintained the sampling resolution at 480$\times$720.

\begin{figure}[t]
    \centering
    \includegraphics[width=0.95\linewidth]{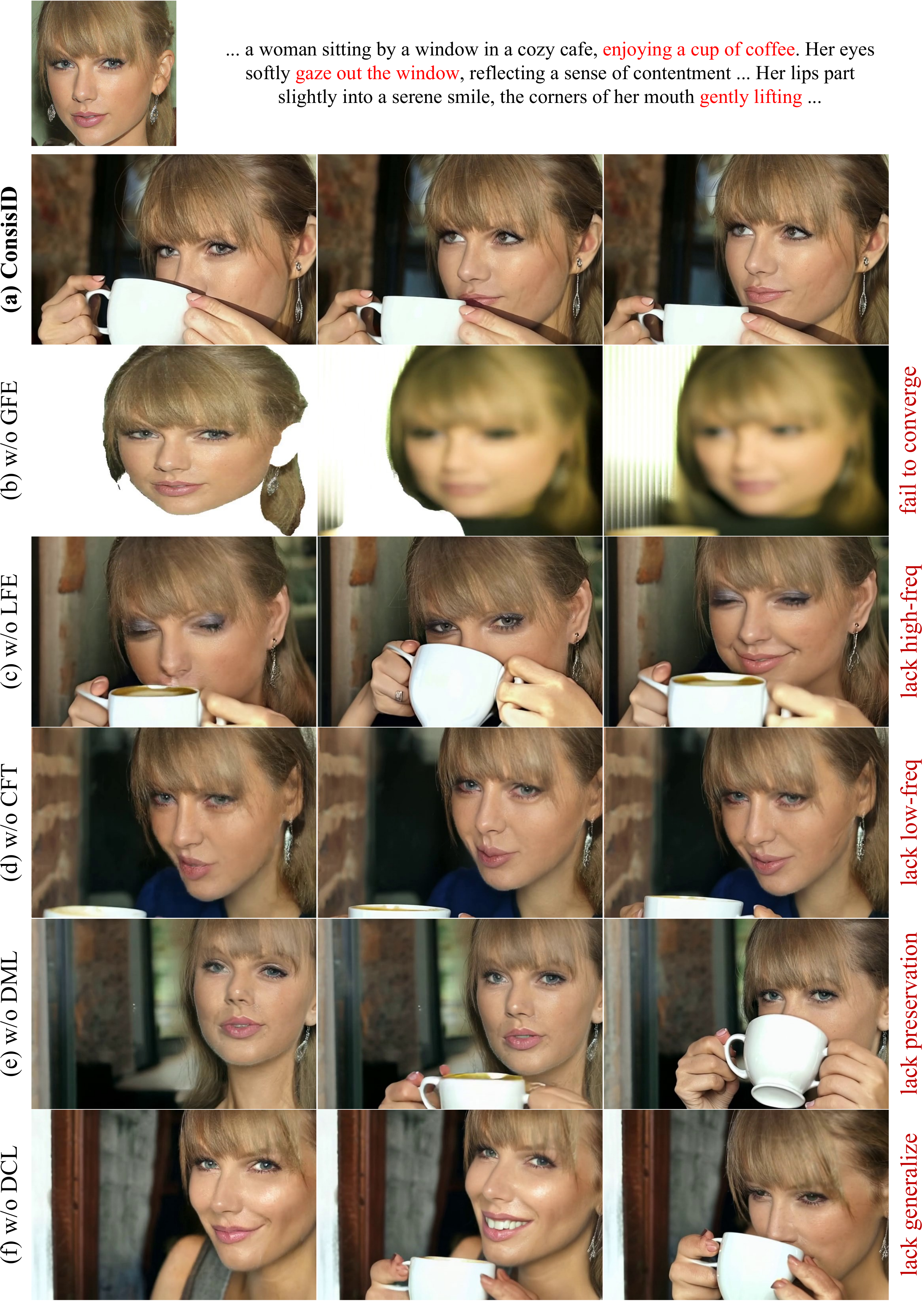}
    \caption{\textbf{Effect of Different Components via Qualitative Analysis.} Removing any component may result in the loss of high- or low-frequency facial information, or hinder the ability to modify video content based on the text prompt.}
    \label{fig: ablation_component}
    % \vspace{-3pt}
\end{figure}

\noindent\myparagraph{Benchmark.}
Since there is an absence of an evaluation dataset, we select 30 persons who were not included in the training data and sourced five high-quality images for each ID from the internet. We then design 90 distinct prompts, encompassing a variety of expressions, actions, and backgrounds for evaluation. Building on previous works \cite{movie_gen, ID-Animator}, we evaluate four dimensions: \textbf{(1)}. Identity Preservation: We use FaceSim-Arc \cite{arcface} and introduce FaceSim-Cur, which assesses identity preservation by measuring feature differences between face regions in the generated videos and those in real face images within the ArcFace \cite{arcface} and CurricularFace \cite{curricularface} feature spaces. \textbf{(2)} Visual Quality: We utilize FID \cite{FID} by calculating feature differences in the face regions between the generated frames and real face images within the InceptionV3 \cite{inceptionv3} feature space. \textbf{(3)} Text Relevance: We utilize CLIPScore \cite{clipscore} to measure the similarity between the generated videos and the input prompts. \textbf{(4)}. Motion Amplitude: Due to the lack of reliable metrics \cite{vbench, chronomagic-bench}, we evaluate through the user study. 
% To reduce overhead, we uniformly sample 16 frames per video to compute the metrics. For fairness, we fix random seeds in all settings.

\subsection{Qualitative Analysis}\label{sec: qualitative analysis}
In this section, we compare our method, ConsisID, with ID-Animator \cite{ID-Animator} (\eg, the only available open-source model) for tuning-free IPT2V tasks. We randomly select images and text prompts of four individuals for qualitative analysis, all of which are absent from the training data. As shown in Figure \ref{fig: qualitative analysis}, ID-Animator cannot generate human body parts beyond the face and is unable to generate complex actions or backgrounds in response to text prompts (\eg, action, attribute, background), which significantly limits its practical application. In addition, the preservation of the identity is inadequate; for example, in case 1, the reference image appears to be processed with skin smoothing. In case 2, wrinkles have been introduced which detract from the aesthetic quality. In cases 3 and 4, the face is distorted due to the lack of low frequency information, which compromises identity consistency. In contrast, the proposed ConsisID consistently produces high-quality, realistic videos that accurately match the reference identity and adhere to prompt.

\begin{figure}[!t]
    \centering
    \includegraphics[width=0.96\linewidth]{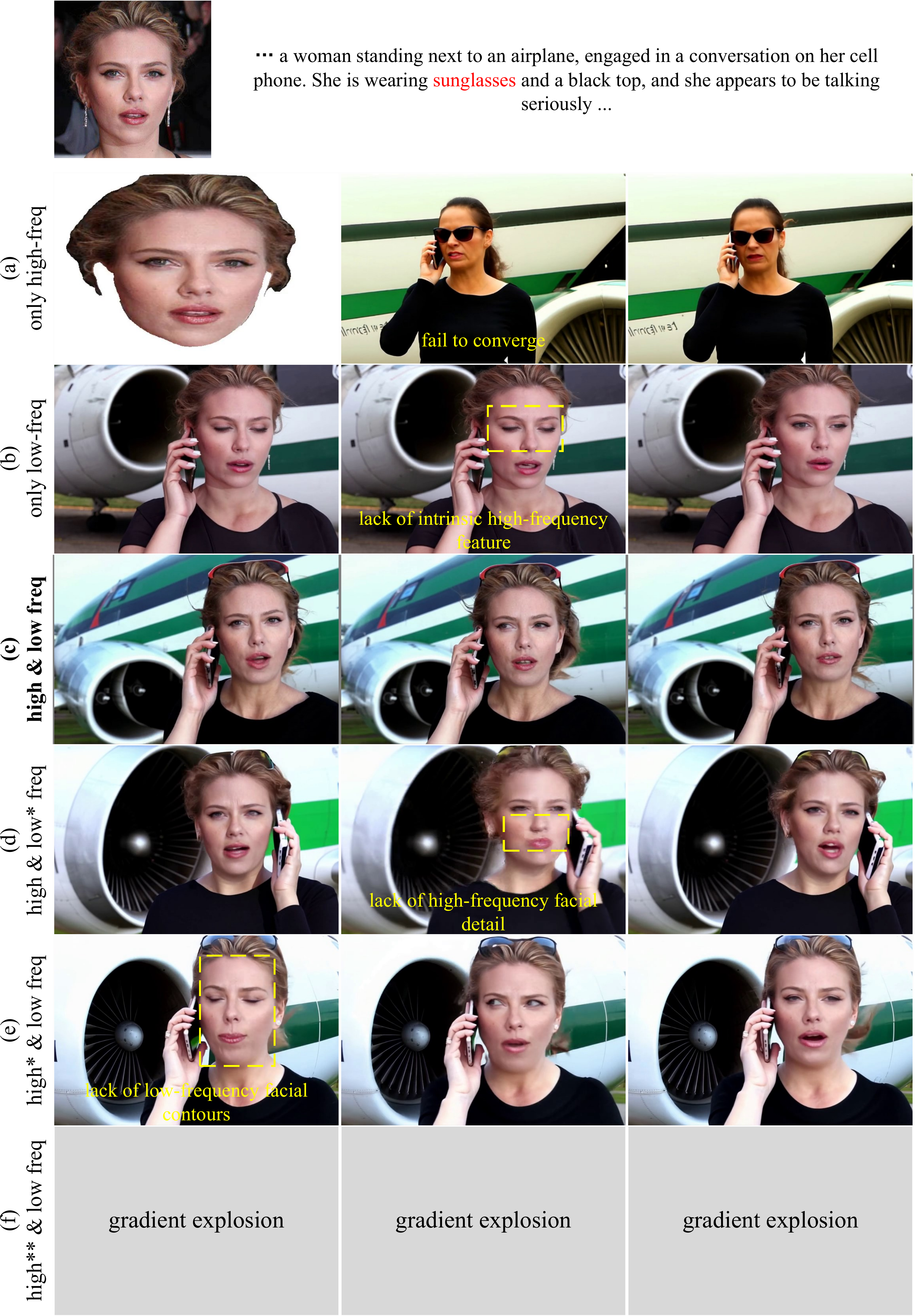}
    \caption{\textbf{Effect of Different Control Signal Injection Way via Qualitative Analysis.} Only (c), which injects both high \& low-freq face signals into the suitable location, performs best.}
    \label{fig: validation_on_id_signal_injection}
    % \vspace{-5pt}
\end{figure}

\subsection{Quantitative Analysis}\label{sec: quantitative analysis}
We present a comprehensive quantitative evaluation of different methods, with results displayed in Table \ref{tab: quantitative analysis}. Consistent with Figure \ref{fig: qualitative analysis}, our method outperforms state-of-the-art methods across five metrics. For identity preservation, ConsisID achieves a higher score by designing appropriate identity signals for DiT from a frequency perspective. By contrast, ID-Animator \cite{ID-Animator} is not optimized for IPT2V and only partially retains facial features, resulting in lower FaceSim-Arc \cite{arcface} and FaceSim-Cur scores. For Text Relevance, ConsisID not only controls expressions via prompts but also adjusts actions and backgrounds, achieving higher CLIPScore \cite{clipscore}. Regarding visual quality, the FID is presented solely as a reference due to its limited alignment \cite{chronomagic-bench, FETV, toward_verifiable, Evalcrafter} with human perception. Please refer to Figure \ref{fig: qualitative analysis} and \ref{fig: human_evaluation} for qualitative analysis of the visual quality.

\subsection{User Study}\label{sec: human evaluation}
Building on previous work, we conduct a human evaluation using a binary voting strategy, with each questionnaire containing only 80 questions. Participants are required to view 40 video clips, a setup designed to improve both engagement and questionnaire validity. For the IPT2V task, each question requires participants to separately judge which option performs better in terms of Identity Preservation, Visual Quality, Text Alignment, and Motion Amplitude. This composition ensures the accuracy of the human evaluation. Owing to the extensive participant base required for this evaluation, we successfully gathered 103 valid questionnaires. The results, depicted in Figure \ref{fig: human_evaluation}, demonstrate a significant superiority of our method over ID-Animator \cite{ID-Animator}, verifying the effectiveness of the designed DiT for IPT2V generation. 

\begin{table}[t]
    \centering
    \resizebox{\columnwidth}{!}
    {
        \begin{tabular}{c|ccccc}
            \toprule
             & FaceSim-Arc $\uparrow$ & FaceSim-Cur $\uparrow$ & CLIPScore $\uparrow$ & FID $\downarrow$ \\
            \midrule
            w/o GFE & 0.05 & 0.05 & 34.86 & 269.88 \\
            w/o LFE & 0.66 & 0.68 & 34.48 & \textbf{104.34} \\
            w/o CFT & 0.54 & 0.58 & 34.47 & 144.62 \\
            w/o DML & 0.62 & 0.67 & 34.23 & 187.78 \\
            w/o DCL & 0.65 & 0.69 & 32.21 & 117.80 \\
          \rowcolor{myblue}  \textbf{ConsisID} & \textbf{0.73} & \textbf{0.75} & \textbf{36.77} & 127.42 \\
            \bottomrule
        \end{tabular}
    }
    \caption{\textbf{Effect of \textit{Local Facial Extractor (LFE)}, \textit{Global Facial Extractor (GFE)}, \textit{coarse-to-fine training (CFT)}, \textit{dynamic mask loss (DML)} and \textit{dynamic cross-face loss (DCL)} by Automatic Metrics.} Removing any of the above methods significantly reduces identity preservation, text relevance, and visual quality.}
    \label{tab: ablation_study_for_quantitative_analysis}
    % \vspace{-5pt}
\end{table}

\subsection{Effect of the Identity Signal Injection in DiT}\label{sec: Effect of the Identity Signal Injection in DiT}
\begin{figure}[t]
    \centering
    \includegraphics[width=0.91\linewidth]{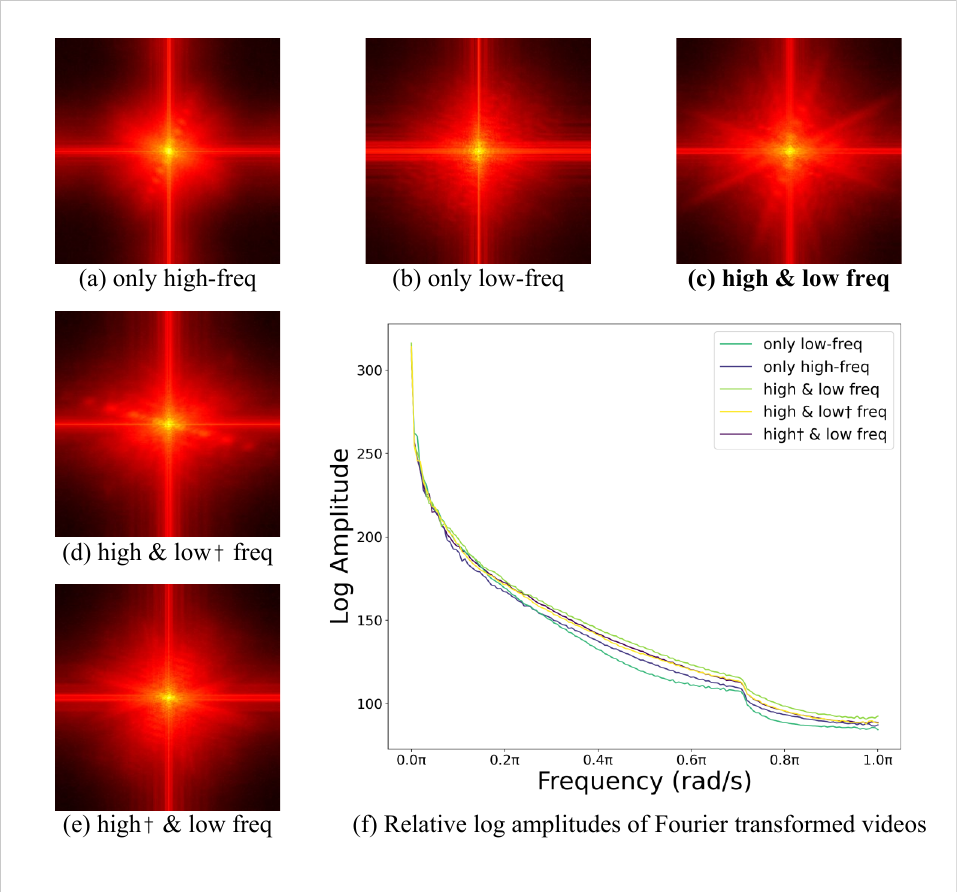}
    \caption{\textbf{(a - e) Fourier spectrum of different id signal injection.} The center area represents low frequencies and the surrounding area represents high frequencies. \textbf{(f) Relative log amplitudes of Fourier transformed generated videos.} A larger response value indicates a higher inclusion of frequency information. (a - f) verify the effect of our frequency decomposition.}
    \label{fig: injection_of_frequency}
    % \vspace{4pt}
\end{figure}

To assess the effectiveness of \textit{Finding 1} and \textit{Finding 2}, we perform ablation studies on different methods of injecting control signals into DiT. Specifically, these experiments involved \textbf{(a)} injecting only low-frequency face information with key points into the noise latent, \textbf{(b)} injecting only high-frequency face signals within the attention block, \textbf{(c)} combining (a) and (b), \textbf{(d)} based on (c), but the low-frequency face information does not contain key points, and \textbf{(e - f)} based on (c), but the high-frequency signal is injected at the output or input of the attention block. \textbf{(g)} injecting only high-frequency face signals before the attention block. To reduce overhead, for each identity, we only select 2 reference images each with 90 text prompts for the evaluation. The results are shown in Figure \ref{fig: validation_on_id_signal_injection} and Table \ref{tab: validation_on_id_signal_injection}. For \textit{Finding 1}, we observe that only injecting high-frequency signals (a) greatly increases the training difficulty, causing the model to fail to converge due to the lack of low-frequency signal injection. In addition, the inclusion of facial key points (d) allows a greater focus on low-frequency information, thereby facilitating training and improving model performance. For \textit{Finding 2}, when only low-frequency signals are injected (b), the model lacks high-frequency information. This reliance on low-frequency signals causes the generated face in the video to copy the reference image, making it difficult to control facial expressions, movements, and other features through prompts. Furthermore, injecting identity signals into the attention block input (f - g) disrupts the intended frequency domain distribution of DiT, resulting in a gradient explosion. Embedding control signals in the attention block (c) is preferable to embedding them in the output (e) because attention block processes predominantly low-frequency information. By embedding high-frequency information internally, the attention block is guided to highlight intrinsic facial features, whereas injecting it into the output merely concatenates features without directing focus, reducing DiT's modeling capacity. Moreover, we apply a Fourier transform to the generated videos (only the face region) to visually compare the influence of different components to extract facial information. As shown in Figure \ref{fig: injection_of_frequency}, the Fourier spectrum and the log amplitude of the Fourier transform reveal that injecting high or low-frequency signals can indeed enhance the corresponding frequency information of the generated face. Moreover, the low-frequency signal can be further enhanced by matching with the face key points, and injecting the high-frequency signal into the attention block has the highest feature utilization rate. Our method (c) shows strongest high and low frequency, further validating the efficiency benefit from \textit{Findings 1 and 2}.

\begin{table}[t]
    \centering
    \resizebox{\columnwidth}{!}
    {
        \begin{tabular}{ccccccc}
            \toprule
            Plan & FaceSim-Arc $\uparrow$ & FaceSim-Cur $\uparrow$ & CLIPScore $\uparrow$ & FID $\downarrow$ \\
            \midrule
            a & 0.05 & 0.05 & 34.86 & 269.88 \\
            b & 0.66 & 0.68 & 34.48 & \textbf{104.34} \\
           \rowcolor{myblue} \textbf{c} & \textbf{0.73} & \textbf{0.75} & \textbf{36.77} & 127.42 \\
            d & 0.64 & 0.68 & 30.69 & 177.65 \\
            e & 0.62 & 0.66 & 33.61 & 164.15 \\
            f & \multicolumn{4}{c}{\textit{unstable training process}} \\
            g & \multicolumn{4}{c}{\textit{unstable training process}} \\
            \bottomrule
        \end{tabular}
    }
    \caption{\textbf{Effect of Different Control Signal Injection Way via Quantitative Analysis.} Only \textbf{plan c}, which injects both high and low-frequency face information into the model, performs best.}
    \label{tab: validation_on_id_signal_injection}
    \vspace{-0.5pt}
\end{table}

\subsection{Ablation on the Consistency Training Strategy}\label{sec: ablation study}
To reduce overhead, for each identity, we only select 2 reference images for the following experiments. To demonstrate the benefits of the proposed consistency training strategy, we perform ablation experiments on coarse-to-fine training (CFT), dynamic mask loss $\mathcal{L}_{e}$ (DML), and dynamic cross-face loss $\mathcal{L}_{f}$ (DCL), with the results presented in Figure \ref{fig: ablation_component} and Table \ref{tab: ablation_study_for_quantitative_analysis}. When CFT is removed, GFE and LFE exhibit competing behaviors, complicating the model's ability to prioritize high and low-frequency information accurately, leading to convergence at suboptimal points. Removing DML required the model to simultaneously focus on both foreground and background elements, with background noise negatively affecting training and reducing facial consistency. Similarly, the exclusion of DCL impaired the generalization capability, reducing fidelity for faces, not in the training set and reducing its effectiveness in generating identity-preserving videos as intended.

% \noindent\myparagraph{Effect of the Consistency Training Strategy.}

\section{Conclusion}\label{sec: conclusion}
In this paper, we present \textbf{ConsisID}, a unified framework for keeping faces consistent in video generation by frequency decomposition. It can seamlessly integrate into existing DiT-based text-to-video models, for generating high-quality, editable, consistent identity-preserving videos. Extensive experiments show that \textbf{ConsisID} outperforms the current state-of-the-art identity-preserving T2V models. It reveals that our frequency-aware heuristic DiT-based control scheme is an optimal solution for IPT2V generation. 

\section{Acknowledgments}
We thank all the anonymous reviewers for their constructive comments. This work was supported in part by the Natural Science Foundation of China (No. 62202014, 62332002, 62425101, 62088102).

% \section{Limitations and Future Work}\label{sec: limitation and future work}
% Existing metrics do not accurately measure the capabilities of different ID preservation models. Although ConsisID can generate realistic and natural videos following a text prompt, metrics such as CLIPScore \cite{clipscore} and FID \cite{FID} show little difference from previous methods. A viable direction is to find a metric that is more in line with human.
\newpage
{
    \small
    \bibliographystyle{ieeenat_fullname}
    \bibliography{main}
}

\clearpage
\setcounter{page}{1}
\setcounter{figure}{0}
\setcounter{table}{0}
\maketitlesupplementary

\startcontents[chapters]
\setcounter{section}{0}
\printcontents[chapters]{}{1}{}

\section{ConsisID Dataset}
\begin{figure*}[!t]
  \centering
  \includegraphics[width=0.9\linewidth]{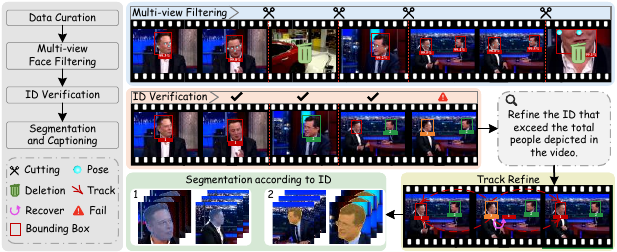}
  \caption{\textbf{Overview of the proposed Identity-Preserving Video Data Processing Pipeline.} First, we identify video clips with high facial density using bounding boxes (bbox) and key points. Next, we implement a tracking algorithm for ID verification, optimizing individual tracking IDs through the previous bounding box. Finally, we generate masks for each individual according to their unique IDs.}
  \label{fig: data_pipeline}
\end{figure*}

We propose a data pipeline to process and construct a high-quality ID-preservation video dataset as shown in Figure \ref{fig: data_pipeline}.

\begin{figure}[!t]
  \centering
  \includegraphics[width=0.90\linewidth]{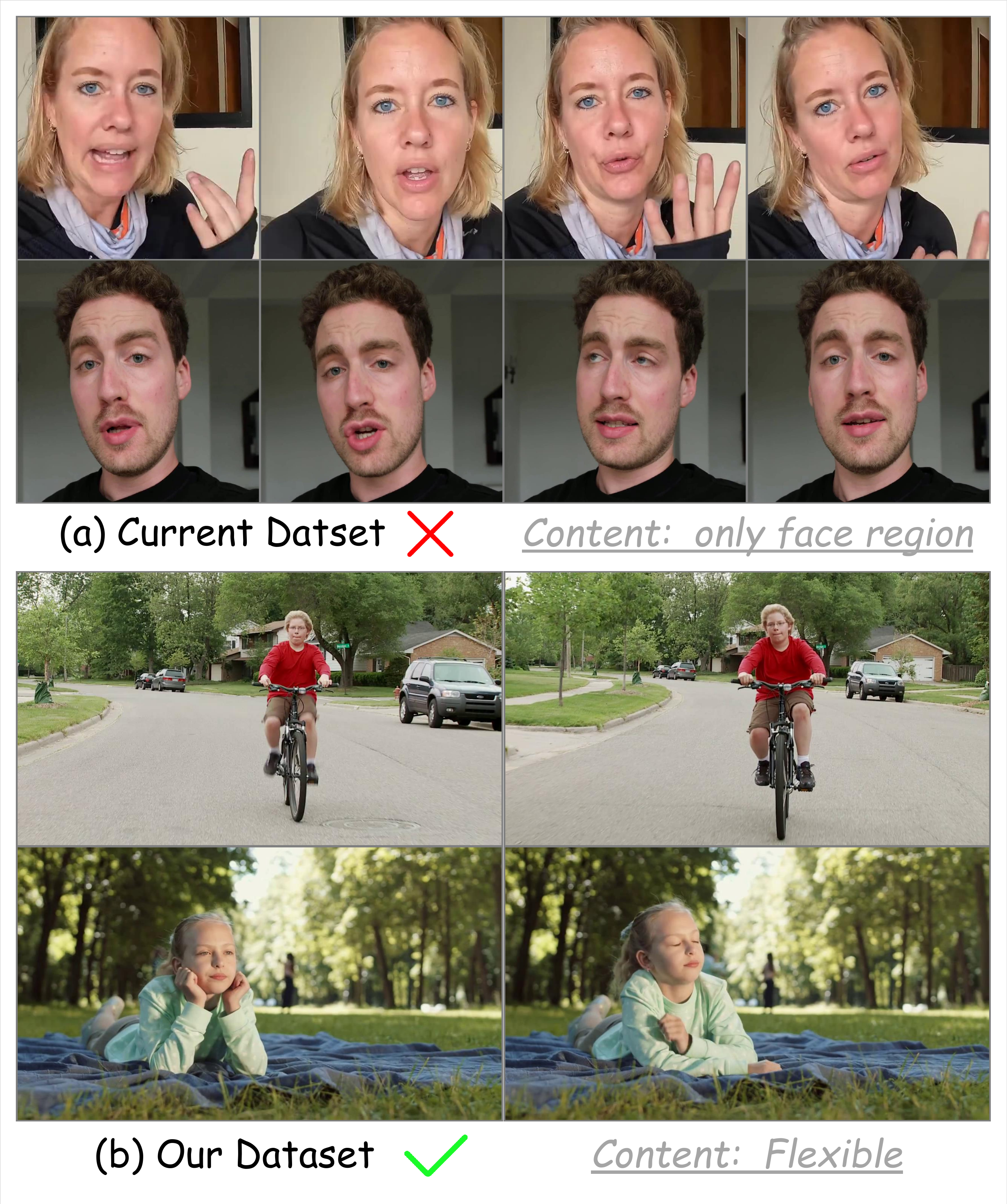}
  \caption{\textbf{Comparison between our in-house data and current Human Centric Video Dataset \cite{celebv-text, voxceleb, ID-Animator}.} Our dataset is more flexible and diverse compared to previous ones.}
  \label{fig: data_comparison}
\end{figure}

\noindent\myparagraph{Data Curation}
Most existing identity-preserving datasets are image-centric \cite{voxceleb, vggface2, Laion-face, celeba-hq}, with only a few focusing on videos \cite{celebv-text, celebv-hq, facevid}. However, these datasets primarily target facial regions, often cropping out relevant background content (\eg, talking heads \cite{HDTF, TalkingHead-1KH}), which limits their broader applicability. To address this, we propose a pipeline to construct identity-preserving videos suitable for daily-life scenarios, as depicted in Figure \ref{fig: data_comparison}. In line with the approach used by \cite{magictime}, we first constructed a set of search keywords (\eg, "human," "woman," "man") and used them to retrieve videos from the internet. During this process, we excluded videos with \textit{few views} or \textit{likes} to ensure quality. Next, following \cite{chronomagic-bench}, we apply PySceneDetect, Aesthetic, and Motion Score to filter out low-quality clips, ultimately getting a dataset of 130K high-quality clips.

\noindent\myparagraph{Multi-view Face Filtering}
The purity of internet-sourced data is typically low, as full videos often include only brief segments featuring facial content. To address this, we first apply YOLO-Box \cite{ultralytics} to extract frame-by-frame bounding box (bbox) information for the categories "face," "head," and "person". Video clips containing all three bounding boxes are retained. We then use YOLO-Pose \cite{ultralytics} to detect facial keypoints (\eg left eye, right eye, left ear, right ear, nose) and filter out video clips with low keypoint density to obtain clean ID-preservation video clips. To mitigate potential YOLO \cite{ultralytics} errors, we set a tolerance threshold $\alpha$. For instance, if $\beta$ frames among frames $0$ to $49$ lack any of the three bounding boxes or valid keypoints, and $\beta < \alpha$, frames $0$ to $49$ are still retained as a complete video clip. To further enhance data quality, we discard video clips in which the "face" bbox occupies less than $6\%$ of the frame.

\begin{figure*}[!t]
    \centering
    \includegraphics[width=0.90\linewidth]{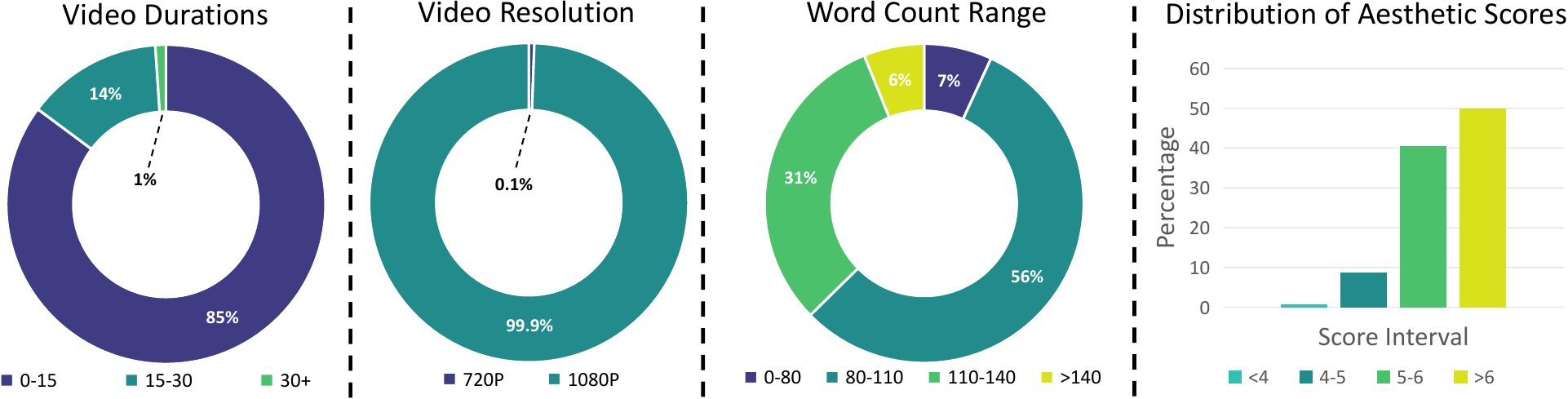}
    \caption{\textbf{Video Clips Statistics of our Dataset.} The dataset includes a diverse range of categories, durations and caption lengths, with most of the videos being in 1080P resolution.
    }
    \label{fig: dataset_static}
\end{figure*}

\noindent\myparagraph{Identical Verification}
A video may include multiple individuals, necessitating the assignment of a unique identifier to each person for subsequent training. Existing video tracking algorithms \cite{BoT-SORT, Bytetrack, Grounded_sam_2} lack robustness, often resulting in missed or incorrect detections. To address this, we utilize the previously obtained frame-by-frame bounding box (bbox) to compute a unique identifier for each individual. Specifically, using the "face" bbox as an example, we first determine the maximum number of individuals, $m$, present in the video based on the bbox information, and then assign a unique identifier (not exceeding $m$) to each "face" bbox in the initial frame. We then apply forward propagation: for the $n$-th frame, each bbox is assigned a unique identifier, corresponds to those from frame $n-1$, based on the Intersection over Union (IoU) between all bboxes of frames $n-1$ and $n$. After completing forward propagation for all frames, we perform backward propagation: for each bbox in frame $n$, we assign a unique identifier that corresponds to frames $n-1$ and $n+1$, again based on the IoU between all bboxes of frames $n$ and $n+1$. Ultimately, each bbox is assigned a unique identifier, facilitating precise tracking of each person's location across video.

\noindent\myparagraph{Segmentation and Captioning}
To facilitate the application of dynamic mask loss, we first input the highest-confidence bounding box (bbox) for each category obtained in the previous step into SAM2 \cite{sam2} to generate the corresponding masks for each person's "face," "head," and "person." Subsequently, SAM2's \cite{sam2} tracking signals are used to further refine the unique identifiers assigned earlier. We then employ Time-Aware Annotation \cite{chronomagic-bench}, leveraging Qwen2-VL-72B \cite{Qwen2-VL}, to produce high-quality captions for the video clips. Data statistics are presented in Figure \ref{fig: dataset_static}.

\begin{figure*}[ht]
    \centering
    \includegraphics[width=0.9\linewidth]{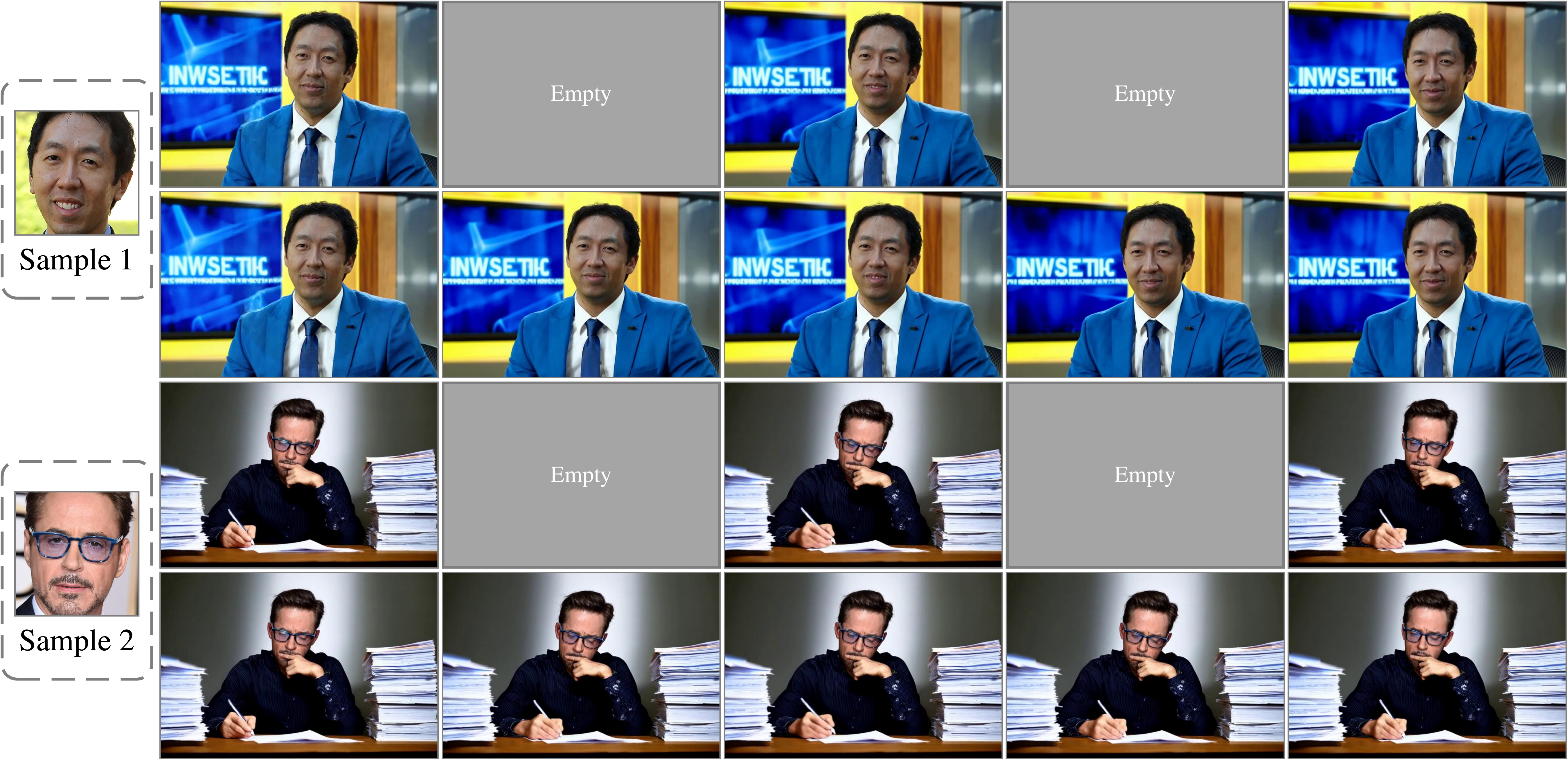}
    \caption{\textbf{Seamlessly integrate the frame interpolation model \cite{rife} into ConsisID.} The newly added frames are clear, which shows that the original video generated by ConsisID is coherent.}
    \label{fig: interpolation}
\end{figure*}

\section{Additional Experimental Results}
\subsection{Comparison with Closed-source Method}\label{sec: Comparison with closed-source model}
In this section, we compare our ConsisID with Vidu 1.5 \cite{Vidu}, the only available video generation model capable of performing the Identity-Preserving Text-to-Video (IPT2V) task. Notably, Vidu 1.5 \cite{Vidu} is a closed-source model, and it remains unclear whether it is tuning-free or tuning-based. As Vidu's API is not publicly available, we can only manually collect output to evaluate its performance. Due to constraints on large-scale generation, we randomly select 60 prompts from the evaluation dataset. Each prompt is paired with a unique reference image, resulting in the generation of 60 videos by each method. The quantitative results are presented in Table \ref{tab: quantitative analysis close-source}, indicating that ConsisID consistently outperforms Vidu 1.5 \cite{Vidu} across all four automatic metrics. The qualitative analysis, illustrated in Figure \ref{fig: comparison_vidu_tuning-base}, further supports these findings. Both ConsisID and Vidu 1.5 follow prompts effectively to generate the specified actions, backgrounds, and attributes. However, Vidu's outputs contain noticeable artifacts (\eg, the flower in case 1 and the tree in case 2) and exhibit lower visual quality compared to ConsisID. Moreover, in terms of preserving identity features, Vidu 1.5 retains only high-frequency facial characteristics (\eg, hair color and hairstyle in case 1, facial shape in case 2), while failing to maintain intrinsic ID features essential for IPT2V. Consequently, individuals generated by Vidu 1.5 do not align consistently with the reference images.

\begin{table}[t]
    \centering
    \resizebox{\columnwidth}{!}
    {
        \begin{tabular}{c|cccc}
            \toprule
             & FaceSim-Arc $\uparrow$ & FaceSim-Cur $\uparrow$ & CLIPScore $\uparrow$ & FID $\downarrow$ \\
            \midrule
            Vidu 1.5 \cite{Vidu} & 0.36 & 0.39 & 32.87 & 215.42 \\
         \rowcolor{myblue}   \textbf{ConsisID} & \textbf{0.52} & \textbf{0.54} & \textbf{33.08} & \textbf{163.68} \\
            \bottomrule
        \end{tabular}
    }
    \caption{\textbf{Quantitative Comparison with Close-source Method.} ConsisID performs best on most metrics, especially FaceSim, which is the most important metric of IPT2V. "$\downarrow$" denotes lower is better. "$\uparrow$" denotes higher is better.}
    \label{tab: quantitative analysis close-source}
\end{table}

\subsection{Comparison with Tuning-based Methods}\label{sec: Comparison with tuning-based methods}
\begin{figure*}[ht]
    \centering
    \includegraphics[width=0.85\linewidth]{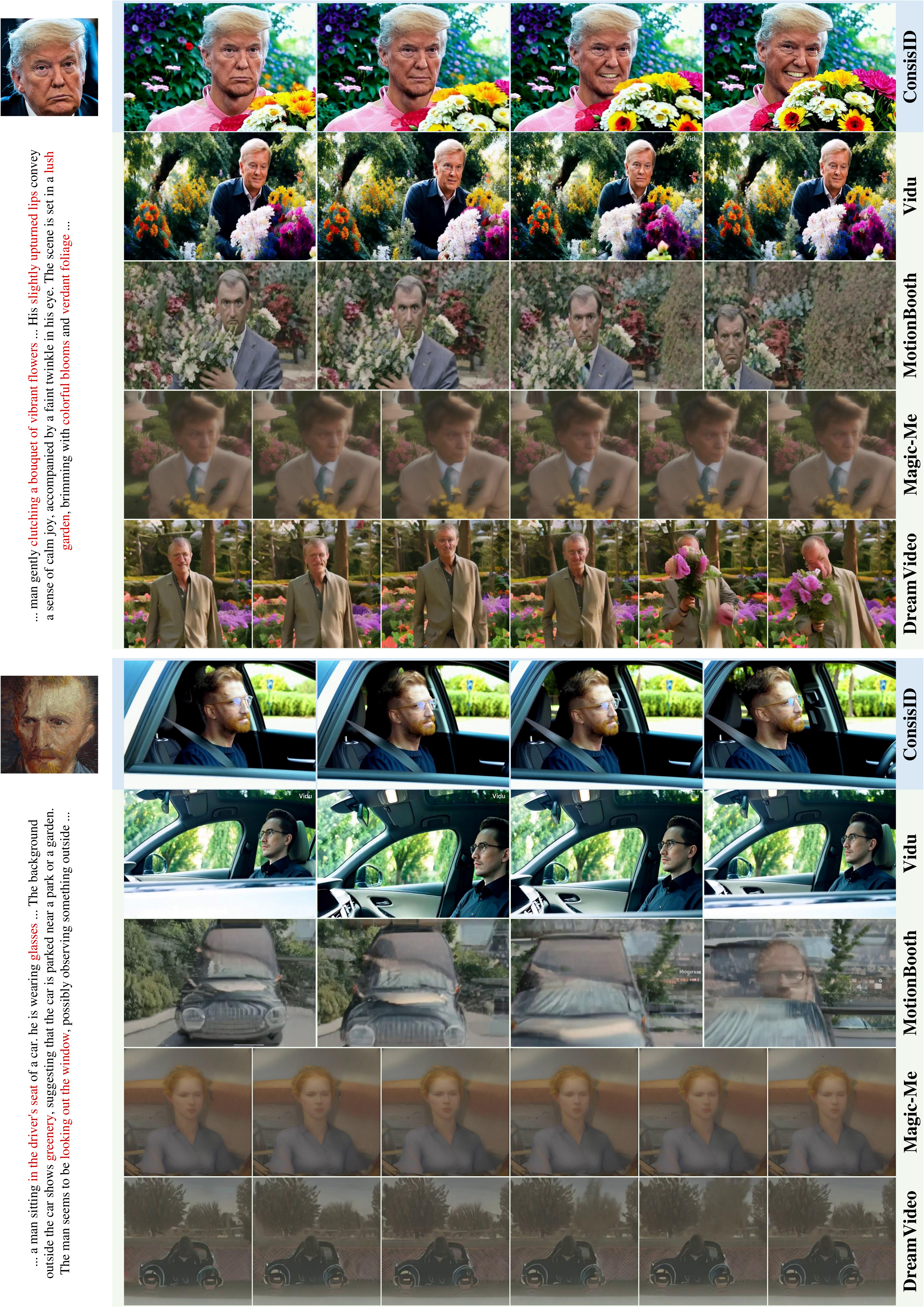}
    \caption{\textbf{Quantitative comparison with Closed-source and Tuning-based Identity-Preserving Videos Generation methods.} ConsisID achieves advantages in identity preservation, visual quality, motion amplitude, and text relevance. Moreover, only ConsisID, Vidu \cite{Vidu} and MotionBooth \cite{motionbooth} can generate aesthetically pleasing \textit{horizontal} videos, while the others \cite{magic-me, dreamvideo} can only produce \textit{square} videos.}
    \label{fig: comparison_vidu_tuning-base}
\end{figure*}

In this section, we compare ConsisID with several tuning-based methods, including DreamVideo \cite{dreamvideo}, MotionBooth \cite{motionbooth} and Magic-Me \cite{magic-me}. These methods require parameter tuning for each new identity input prior to inference. Due to computational and time constraints, we randomly select 1 reference image per ID from our evaluation dataset and each image is evaluated on 45 randomly selected prompts, resulting in a total of 1,350 video sequences generated per method. The results are presented in Figure \ref{fig: comparison_vidu_tuning-base} and Table \ref{tab: quantitative analysis tuning-free}. As shown, ConsisID consistently outperforms existing tuning-based methods despite having a shorter inference time (on single Nvidia H100). This superior performance is likely due to the fact that the latter are designed for open-domain tasks (\eg, people, objects, animals, plants), which encompass a wide range and consequently fail to effectively capture the nuanced distinctions of individual identity features, leading to suboptimal results.

\begin{table}[!t]
    \centering
    \resizebox{\columnwidth}{!}
    {
        \begin{tabular}{c|ccccc}
            \toprule
             & FaceSim-Arc $\uparrow$ & FaceSim-Cur $\uparrow$ & CLIPScore $\uparrow$ & FID $\downarrow$ & Speed (s) $\downarrow$ \\
            \midrule
            DreamVideo \cite{dreamvideo} & 0.03 & 0.03 & 26.03 & 237.91 & 3600+ \\
            MotionBooth \cite{motionbooth} & 0.05 & 0.06 & 24.42 & 287.90 & 600+ \\
            Magic-Me \cite{magic-me} & 0.09 & 0.10 & 23.14 & 237.35 & 500+ \\
         \rowcolor{myblue}   \textbf{ConsisID} & \textbf{0.46} & \textbf{0.47} & \textbf{27.45} & \textbf{181.97} & 100+ \\
            \bottomrule
        \end{tabular}
    }
    \caption{\textbf{Quantitative Comparison with Tuning-based Methods (on single Nvidia H100).} ConsisID can generate high-quality id-preserving videos in a very short time, achieving the best balance among methods. "$\downarrow$" denotes lower is better. "$\uparrow$" higher is better.}
    \label{tab: quantitative analysis tuning-free}
\end{table}

\begin{table}[!t]
    \centering
    \resizebox{\columnwidth}{!}
    {
        \begin{tabular}{c|cccc}
            \toprule
             & FaceSim-Arc $\uparrow$ & FaceSim-Cur $\uparrow$ & CLIPScore $\uparrow$ & FID $\downarrow$  \\
            \midrule
            CogVideoX-5B-I2V \cite{cogvideox} & 0.37 & 0.38 & \textbf{28.53} & 201.69  \\
            EasyAnimate v4 \cite{easyanimate} & 0.15 & 0.15 & 27.95 & 235.68 \\
            OpenSora-Plan v1.3 \cite{opensoraplan} & 0.31 & 0.32 & 27.25 & 224.99 \\
            DynamiCrafter512 \cite{Dynamicrafter} & 0.25 & 0.26 & 29.76 & 212.13 \\
            \rowcolor{myblue}   \textbf{ConsisID} & \textbf{0.46} & \textbf{0.47} & 27.45 & \textbf{181.97} \\
            \bottomrule
        \end{tabular}
    }
    \caption{\textbf{Quantitative Comparison with I2V Methods.} End-to-end methods yield higher-quality id-preserving videos than two-stage methods. "$\downarrow$" denotes lower is better. "$\uparrow$" higher is better.}
    \label{tab: Quantitative comparison with I2V Methods}
\end{table}

\subsection{Comparison with I2V Methods}\label{sec: Comparison with I2V Methods}
\begin{figure}[!t]
    \centering
    \includegraphics[width=1\linewidth]{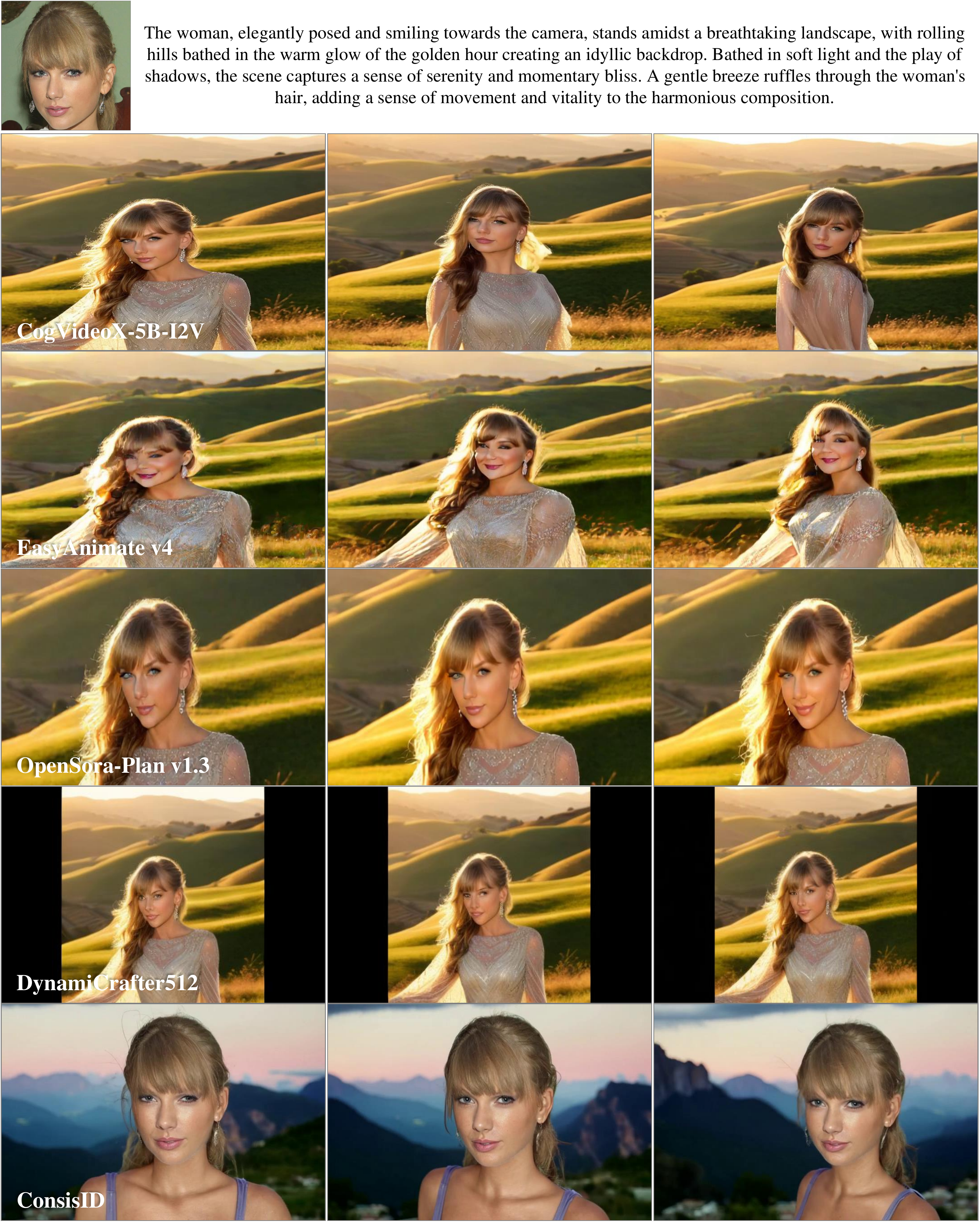}
    \caption{\textbf{Qualitative comparison with I2V Methods.} It is clear that standard I2V models encounter challenges in generating high-quality identity-preserving videos.}
    \label{fig: Qualitative comparison with I2V Methods.}
\end{figure}

In this section, we compare ConsisID with several image-to-video methods, leveraging the identity-preserving image model \cite{photomaker}. As shown in Figure \ref{fig: Qualitative comparison with I2V Methods.}, Table \ref{tab: Quantitative comparison with I2V Methods} and Table \ref{tab: Computation Overhead}, the I2V foundation models \cite{cogvideox, easyanimate, Dynamicrafter} clearly demonstrates considerable temporal decay in identity preservation. While OpenSora-Plan \cite{opensoraplan} achieves higher fidelity due to its lower motion amplitude, it does not align with the real video. In contrast, only the proposed ConsisID consistently preserves identity throughout the entire video.

\begin{table}[!t]
    \centering
    \resizebox{\columnwidth}{!}
    {
        \begin{tabular}{c|cccc}
            \toprule
             & FaceSim-Arc $\uparrow$ & FaceSim-Cur $\uparrow$ & CLIPScore $\uparrow$ & FID $\downarrow$  \\
            \midrule
            w/o CLIP & 0.40 & 0.38 & 26.87 & 142.23 \\
            w/o FaceExtractor & 0.36 & 0.37 & 27.76 & 193.99 \\
            \midrule
            w/o Noise $\zeta$ & 0.37 & 0.38 & 27.52 & 193.46 \\
            \midrule
            Loss \dag & 0.35 & 0.37 & 26.53 & 167.11 \\
            Loss \dag\dag & 0.39 & 0.38 & 26.89 & 216.69 \\
            Loss \ddag & 0.29 & 0.30 & 24.77 & 150.80 \\
            \midrule
            ConsisID \dag & 0.40 & 0.40 & 27.38 & 256.29 \\
            \rowcolor{myblue} \textbf{ConsisID} & \textbf{0.46} & \textbf{0.47} & \textbf{27.45} & \textbf{181.97} \\
            \bottomrule
        \end{tabular}
    }
    \caption{\textbf{Fine-grained Ablation Study.} Each component of ConsisID plays a crucial role in generating high-quality videos. "$\downarrow$" denotes lower is better. "$\uparrow$" higher is better.}
    \label{tab: Fine-grained Ablation Study}
\end{table}

\begin{figure}[!t]
    \centering
    \includegraphics[width=1\linewidth]{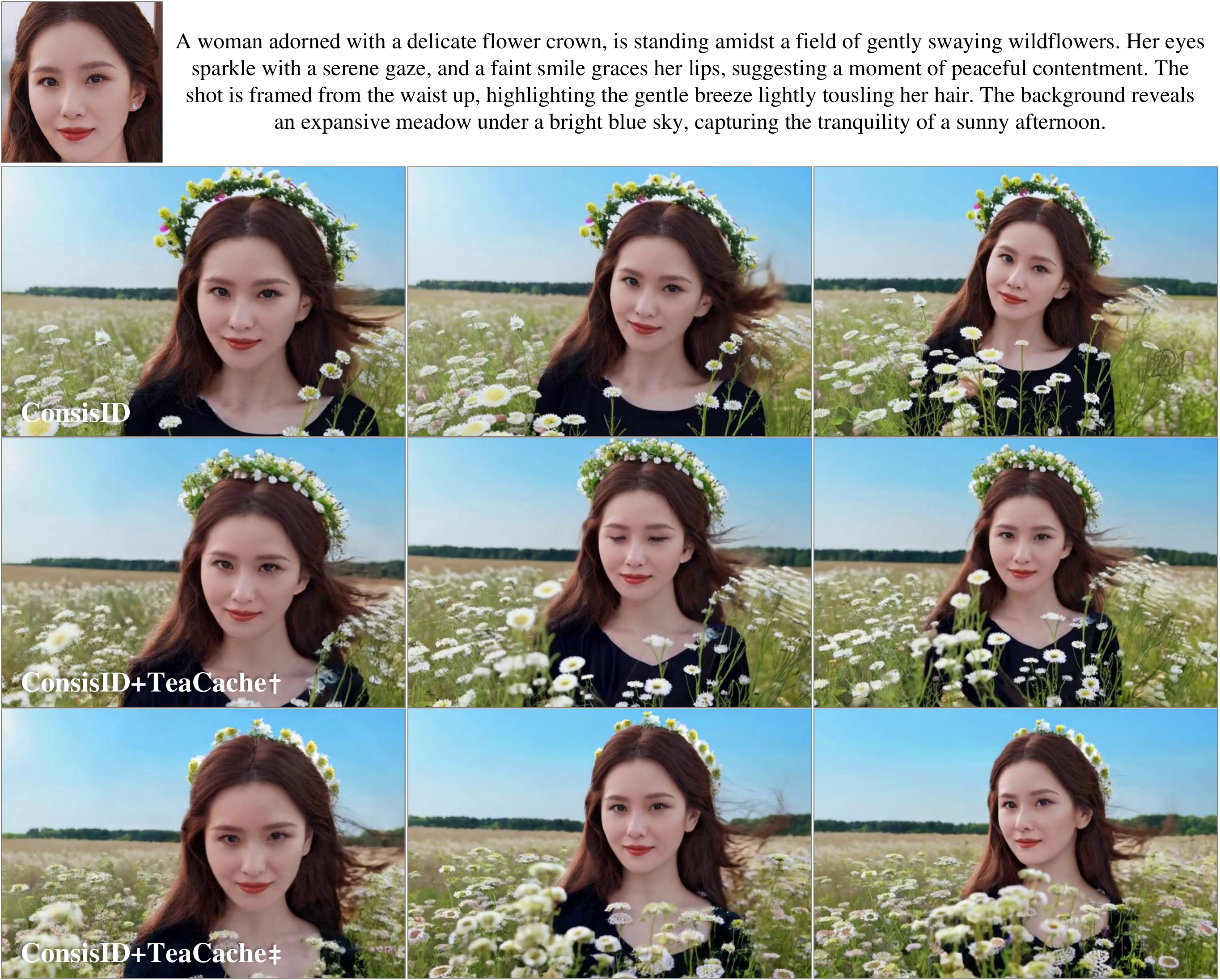}
    \caption{\textbf{Increasing Inference Speed with the help of TeaCache \cite{TeaCache}}. ConsisID can seamlessly integrate into existing inference acceleration frameworks without much quality degradation, demonstrating its strong scalability.}
    \label{fig: visualization_of_teacache}
\end{figure}

\begin{figure*}[!t]
    \centering
    \includegraphics[width=0.9\linewidth]{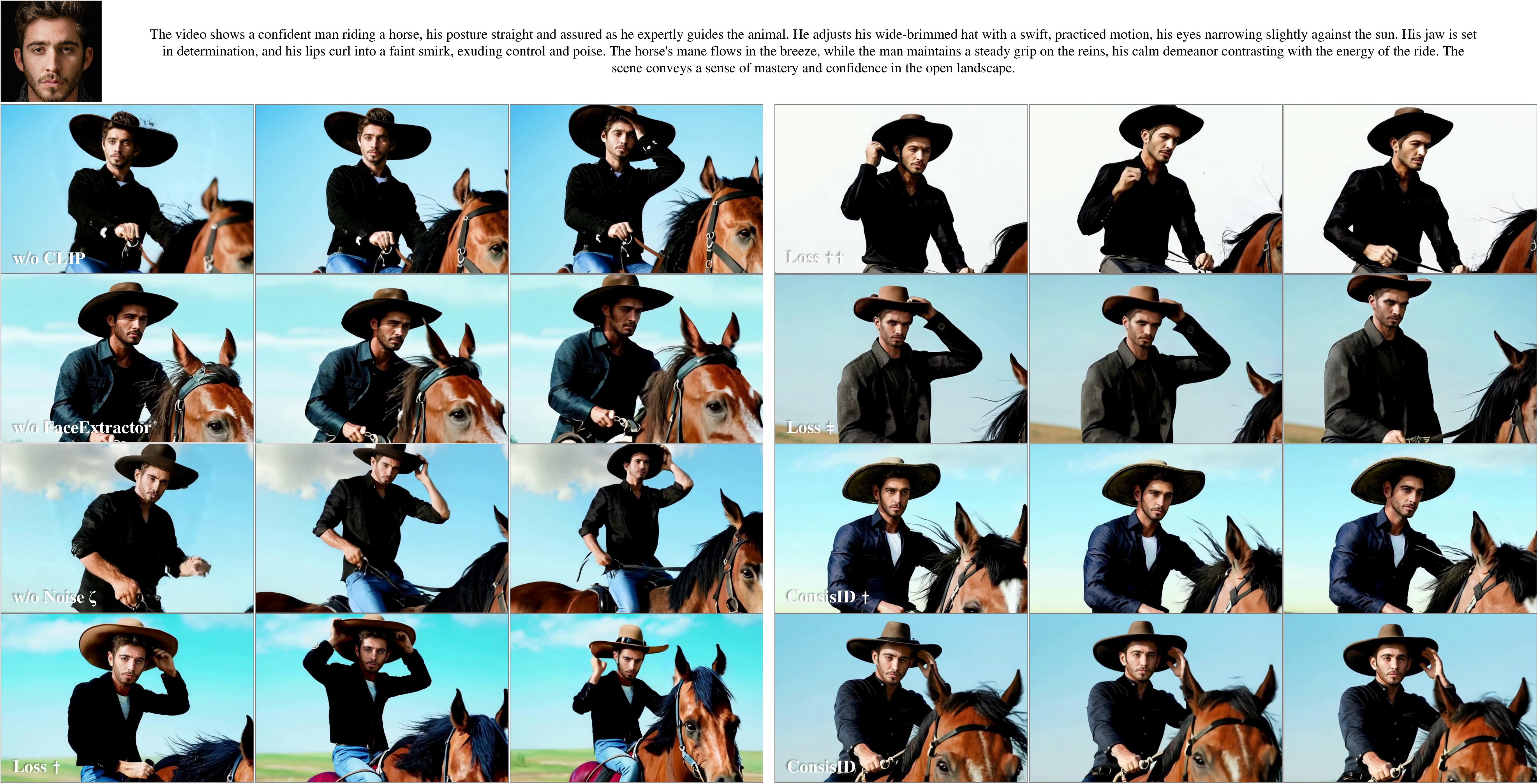}
    \caption{\textbf{Fine-grained Ablation Study.} After the removal of CLIP, the model loses essential semantic information necessary for editing. FaceExtractor and MSE Loss play a critical role in maintaining consistency of facial features. Over-reliance on Dynamic Mask Loss may result in the loss of background information. Noise $\zeta$ and Dynamic Cross Loss are vital for the model's generalization; without them, the model struggles to produce results beyond the training data.}
    \label{fig: Fine-grained Ablation Study}
\end{figure*}

% \subsection{Ablation on Noise in Cross-Face Loss}\label{sec: Ablation on Noise in Cross-Face Loss}
% Pass

% \subsection{Ablation on the Weight Ratio of Loss}\label{sec: Ablation on the Weight Ratio of Loss}
% Pass

% \subsection{Ablation on the Local Facial Extractor}\label{sec: Ablation on the Local Facial Extractor}
% Pass

\subsection{Fine-grained Ablation Study}\label{sec: Fine-grained Ablation Study}
In this section, we conduct more detailed ablation studies. Due to limited computational resources, we extract approximately 30K video samples from the ConsisID-Dataset for the following experiments. The batch size is reduced to 16, and the training duration is set to one epoch, with all other settings remaining unchanged. The experiments including:

\noindent \myparagraph{\textbf{Ablation on the Local Facial Extractor}}: This experiment aims to assess the distinct roles of CLIP and the Face Extractor in capturing high-frequency facial features, specifically \textit{w/o CLIP} and \textit{w/o Face Extractor}.

\noindent  \myparagraph{\textbf{Ablation on Noise in Cross-Face Loss}}: This experiment examines whether introducing subtle noise into the input image improves the model's generalization capability, specifically \textit{w/o Noise $\zeta$}.

\noindent \myparagraph{\textbf{Ablation on the Weight Ratio of Loss}}: This experiment investigates the individual roles of MSE Loss, Dynamic Mask Loss, and Dynamic Cross Loss to the training process. \textit{Loss \dag} corresponds to a mix ratio of 2:1:1, \textit{Loss  \dag\dag} to a ratio of 1:2:1, and \textit{Loss \ddag} to a ratio of 1:1:2.

The results are shown in Figure \ref{fig: Fine-grained Ablation Study} and Table \ref{tab: Fine-grained Ablation Study}, where \textit{ConsisID \dag} represents the complete model trained on the subset data. It can be concluded that CLIP is essential for acquiring the semantic information necessary for video editing, as removing it leads to a significant decrease in the CLIPScore. FaceExtractor plays a critical role in maintaining facial consistency, with its removal resulting in a drop in FaceSim scores. Both Noise $\zeta$ and Dynamic Cross Loss contribute positively to the model's generalization performance; however, an overemphasis on the latter may prevent convergence. MSE Loss accelerates convergence, while Dynamic Mask Loss enhances focus on facial features, thereby improving identity consistency. However, excessive reliance on Dynamic Mask Loss may lose the ability to generate background content. The complete model, integrating all components, yields optimal performance.

\begin{table}[t]
    \centering
    \resizebox{\columnwidth}{!}
    {
        \begin{tabular}{c|ccc}
            \toprule
            Model & Memory & Paramaters & Speed \\
            \midrule
            ID-Animator \cite{ID-Animator} & 8GB & 1.5B & \textasciitilde 11s \\
            CogVideoX-5B-I2V \dag\dag\ \cite{cogvideox} & 42GB & 5.2B & \textasciitilde 210s+7s \\
            EasyAnimate v4 \dag\dag\ \cite{easyanimate} & 15GB & 1.8B & \textasciitilde 78s+7s \\
            OpenSora-Plan v1.3 \dag\dag\ \cite{opensoraplan} & 37GB & 2.6B & \textasciitilde 282s+7s \\
            DynamiCrafter \dag\dag\ \cite{Dynamicrafter} & 17GB & 2.4B & \textasciitilde 26s+7s \\
            \midrule
            CogVideoX-5B-I2V \cite{cogvideox} & 42GB & 5.2B & \textasciitilde 210s \\
            \rowcolor{myblue} \textbf{ConsisID} & 44GB & 5.7B & \textasciitilde 214s \\
            \midrule
            ConsisID+TeaCache \dag\ \cite{TeaCache} & 44GB & 5.7B & \textasciitilde 137s \\
            ConsisID+TeaCache \ddag\ \cite{TeaCache} & 44GB & 5.7B & \textasciitilde 103s \\
            \bottomrule
        \end{tabular}
    }
    \caption{\textbf{Computation Overhead of Different Methods (on single Nvidia A100).} Compared to the baseline, ConsisID introduces only a minimal overhead to achieve the IPT2V task. \dag\dag\ means generating video with the help of PhotomakerV2 \cite{photomaker}.}
    \label{tab: Computation Overhead}
\end{table}

\begin{figure}[!t]
    \centering
    \includegraphics[width=1\linewidth]{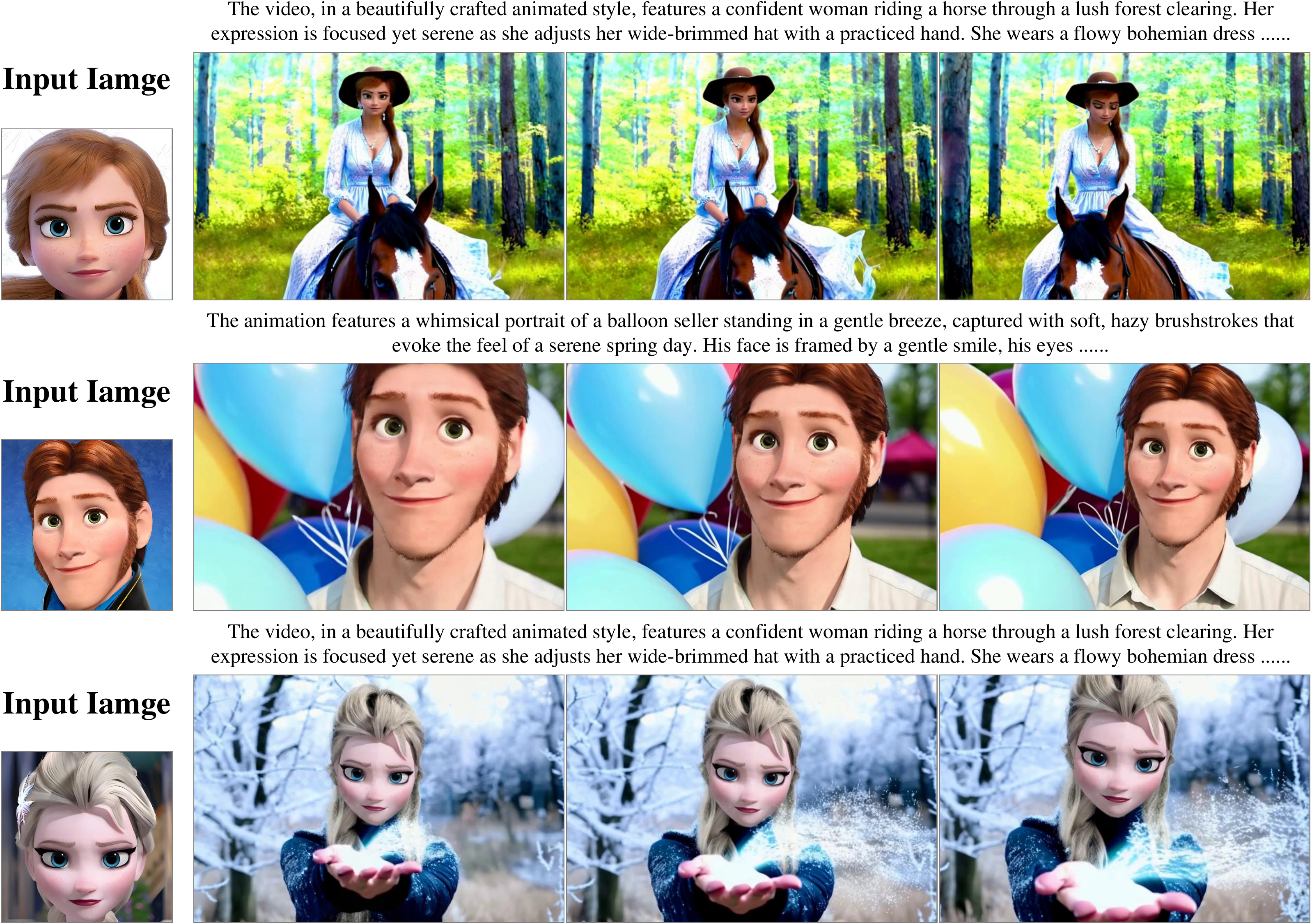}
    \caption{\textbf{Style Transfer Applications.} Despite being trained on real facial data, ConsisID demonstrates a remarkable generalization by generating anime-style videos in a zero-shot manner.}
    \label{fig: style_transfer}
\end{figure}

\subsection{Ablation on the Number of Inference Steps}\label{sec: Effect of the Number of Inference Steps}
\begin{figure*}[!t]
    \centering
    \includegraphics[width=0.90\linewidth]{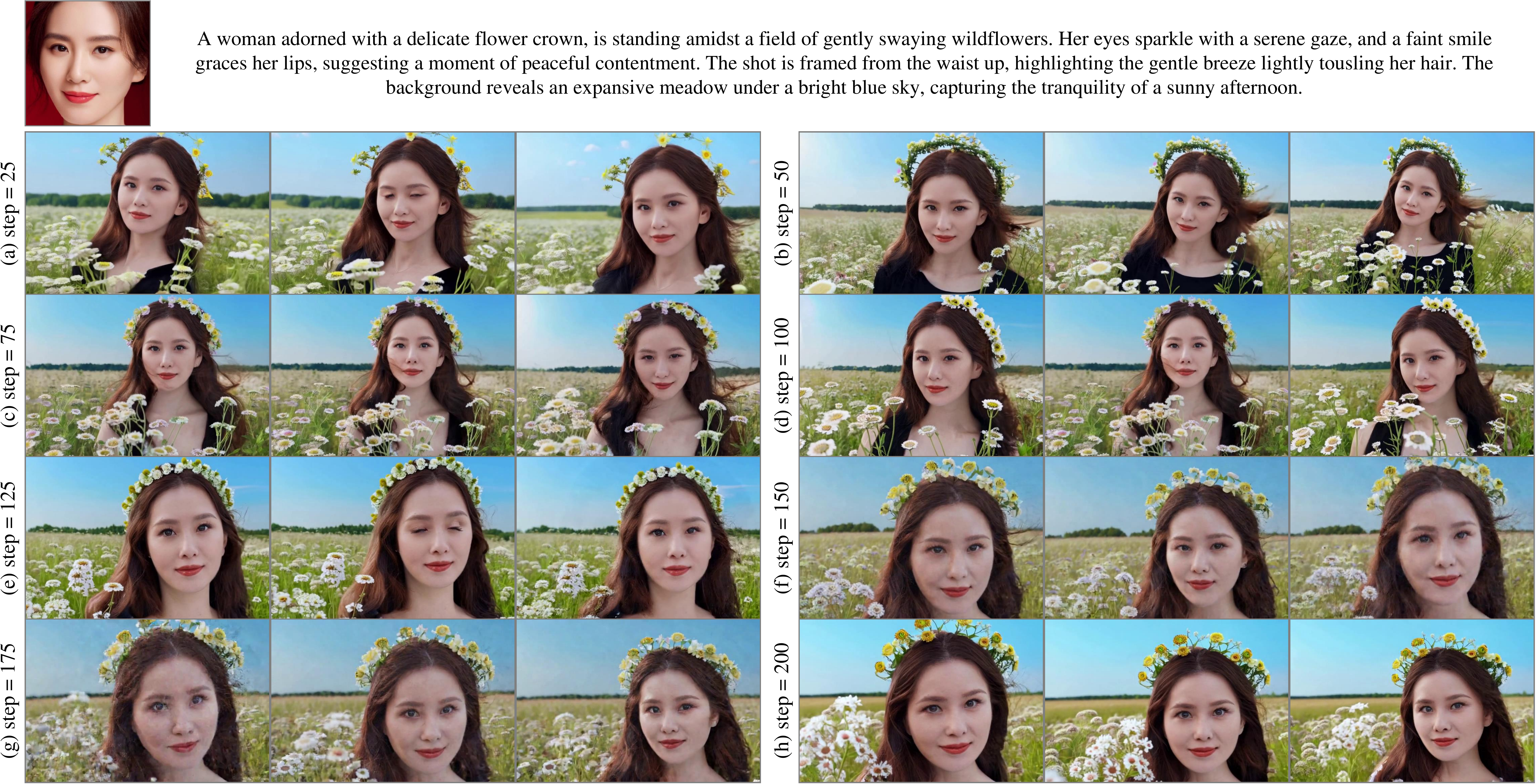}
    \caption{\textbf{Effect of the Inference Steps $t$.} Overall quality does not improve consistently as $t$ increases, but first improves and then declines. This may be because the early steps are dominated by low frequency, whereas the later steps are dominated by high frequency.}
    \label{fig: ablation_step}
\end{figure*}

To assess the impact of varying the number of inference steps on model performance, we conduct an ablation study within the inference phase of ConsisID. Given constraints on computing resources, 60 prompts are randomly selected from the evaluation dataset. Each prompt is paired with a unique reference image, leading to the generation of 60 videos for each setting. Using a fixed random seed, we vary the inversion step parameter \( t \) across values of 25, 50, 75, 100, 125, 150, 175, and 200. The results are illustrated in Figure \ref{fig: ablation_step} and Table \ref{tab:ablation_step}. Although theoretical expectations \cite{DPM, DDIM, DDPM} suggest that increasing the number of inference steps would continuously enhance the generation quality, our findings indicate a non-linear relationship where quality peaks at $t = 50$ and subsequently declines. Specifically, at $t = 25$, the model produces incomplete garlands; at $t = 75$, it fails to generate upper body clothing; beyond $t = 125$, it loses critical low-frequency facial information, resulting in distorted facial features; and beyond $t = 150$, the visual clarity progressively deteriorates. We infer that the initial stages of denoising process are dominated by low-frequency information, such as generating the outline of a face, while the later stages focus on high-frequency details, such as intrinsic facial features. $t = 50$ is just the optimal setting to balance these two stages.

\begin{table}[t]
    \centering
    \resizebox{\columnwidth}{!}{
        \begin{tabular}{c|ccccc}
            \toprule
            & FaceSim-Arc $\uparrow$ & FaceSim-Cur $\uparrow$ & CLIPScore $\uparrow$ & FID $\downarrow$ & Speed (s) $\downarrow$ \\
            \midrule
            $t = 25$  & 0.50 & 0.53 & 30.43 & 184.44 & \textbf{50+} \\
            \rowcolor{myblue} \textbf{$t = 50$}  & \textbf{0.52} & 0.54 & \textbf{33.08} & \textbf{163.68} & 100+ \\
            $t = 75$  & 0.43 & 0.52 & 31.92 & 200.86 & 160+ \\
            $t = 100$ & 0.46 & \textbf{0.55} & 32.25 & 212.74 & 220+ \\
            $t = 125$ & 0.42 & 0.51 & 32.38 & 185.85 & 270+ \\
            $t = 150$ & 0.34 & 0.40 & 32.41 & 186.56 & 330+ \\
            $t = 175$ & 0.35 & 0.42 & 29.98 & 186.99 & 390+ \\
            $t = 200$ & 0.33 & 0.39 & 31.18 & 166.79 & 440+ \\
            \bottomrule
        \end{tabular}
    }
    \caption{\textbf{Effect of the Inference Steps by Quantitative Analysis (on Nvidia H100).} "$\downarrow$" denotes lower is better. "$\uparrow$" higher is better.}
    \label{tab:ablation_step}
    \vspace{-3pt}
\end{table}

\subsection{Increasing Inference Speed}\label{sec: Increasing Inference Speed}
As shown in Table \ref{tab: Computation Overhead}, ConsisID requires about 44 GB of GPU memory to decode 49 frames with output resolution 720x480 ($W \times H$), while baseline needs 42 GB of GPU memory. The inference time of ConsisID is almost identical to the baseline, with only an additional 0.5B parameters, yet it achieves the IPT2V functionality that the baseline cannot, demonstrating the efficiency of the proposed method. In addition, ConsisID can seamlessly integrate with training-free inference acceleration strategies, achieving minimal degradation in visual quality, as illustrated in Figure \ref{fig: visualization_of_teacache}. Specifically, TeaCache\dag\ corresponds to setting $rel\_l1\_thresh = 0.1$, while TeaCache\ddag\ corresponds to setting $rel\_l1\_thresh = 0.15$. The $rel\_l1\_thresh$ regulates the trade-off between generation quality and speed.

\subsection{Generating Higher FPS Videos}\label{sec: Higher Frames Per Second}
Due to limited resources, ConsisID can only generate 49 frames at 8 frames per second (fps). Although the resulting video is coherent, the frame rate falls below the human perceptual threshold for smoothness, which is approximately 16 fps. Therefore, a frame interpolation model \cite{rife} can be applied to post-process the output video, increasing the frame rate to 16 fps, as illustrated in Figure \ref{fig: interpolation}. The results indicate that after interpolation, the video maintains a high level of clarity, suggesting that the original frames generated at fps 8 are sufficiently coherent.

\subsection{Style Transfer Applications}\label{sec: Style Transfer Applications}
Figure \ref{fig: style_transfer} demonstrates the generalization capability of ConsisID. Beyond generating realistic, customized videos, the framework effectively processes stylized prompts while preserving the identity of animated characters in a zero-shot manner. These capabilities are expected to significantly advance video content creation.

\subsection{More Cases on ID-preserving Videos}\label{sec: more cases on personalized videos}
\begin{figure*}[ht]
    \centering
    \includegraphics[width=0.93\linewidth]{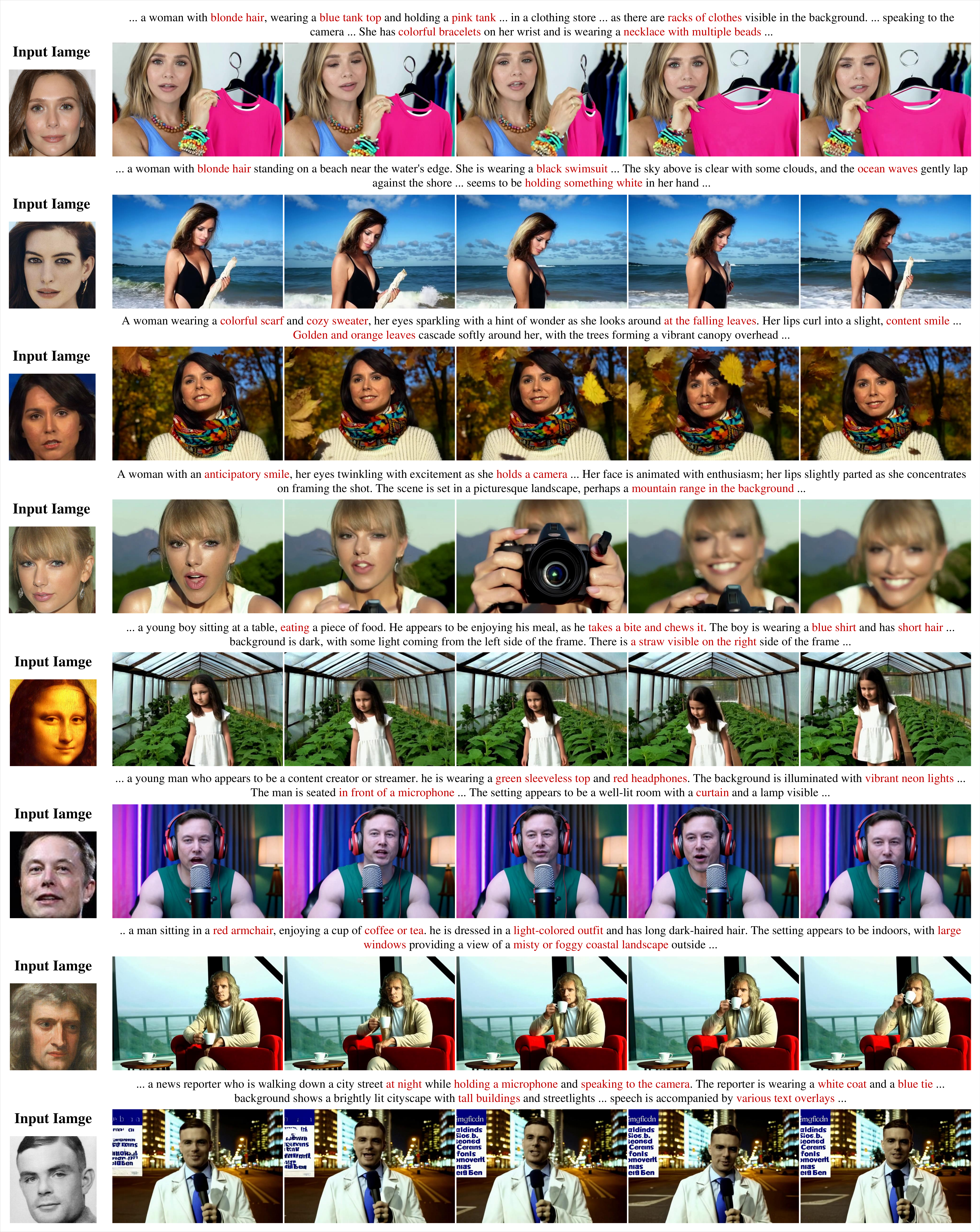}
    \caption{\textbf{Showcases of Identity-Preserving Videos Generated by ConsisID.} Our method consistently generates realistic videos that match the input identity while enabling precise control through text prompts, demonstrating significant practical utility.}
    \label{fig: main results 2}
\end{figure*}

\begin{figure*}[ht]
    \centering
    \includegraphics[width=0.93\linewidth]{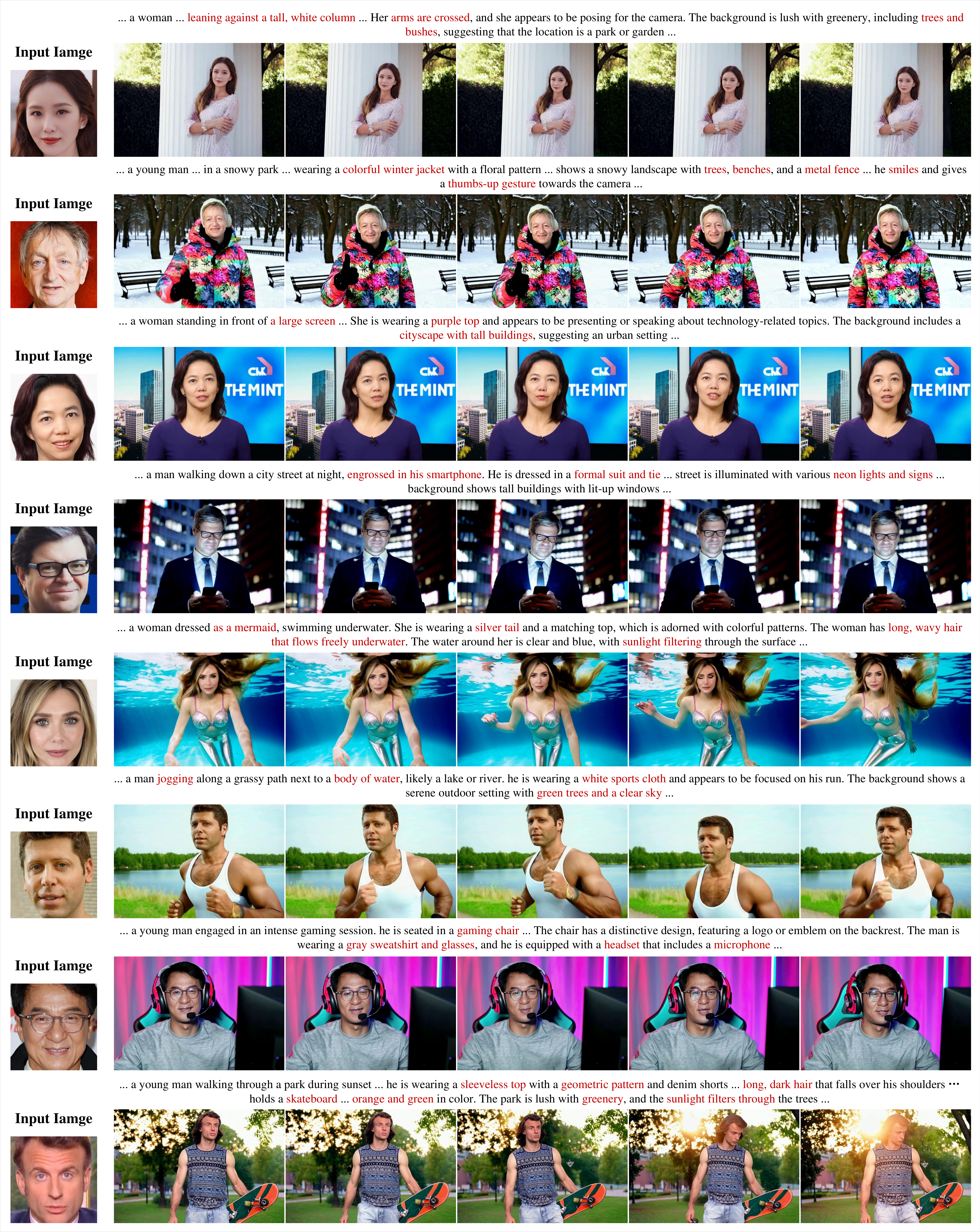}
    \caption{\textbf{More Showcases of Identity-Preserving Videos Generated by ConsisID.}}
    \label{fig: main results 3}
\end{figure*}

\begin{figure*}[ht]
    \centering
    \includegraphics[width=0.93\linewidth]{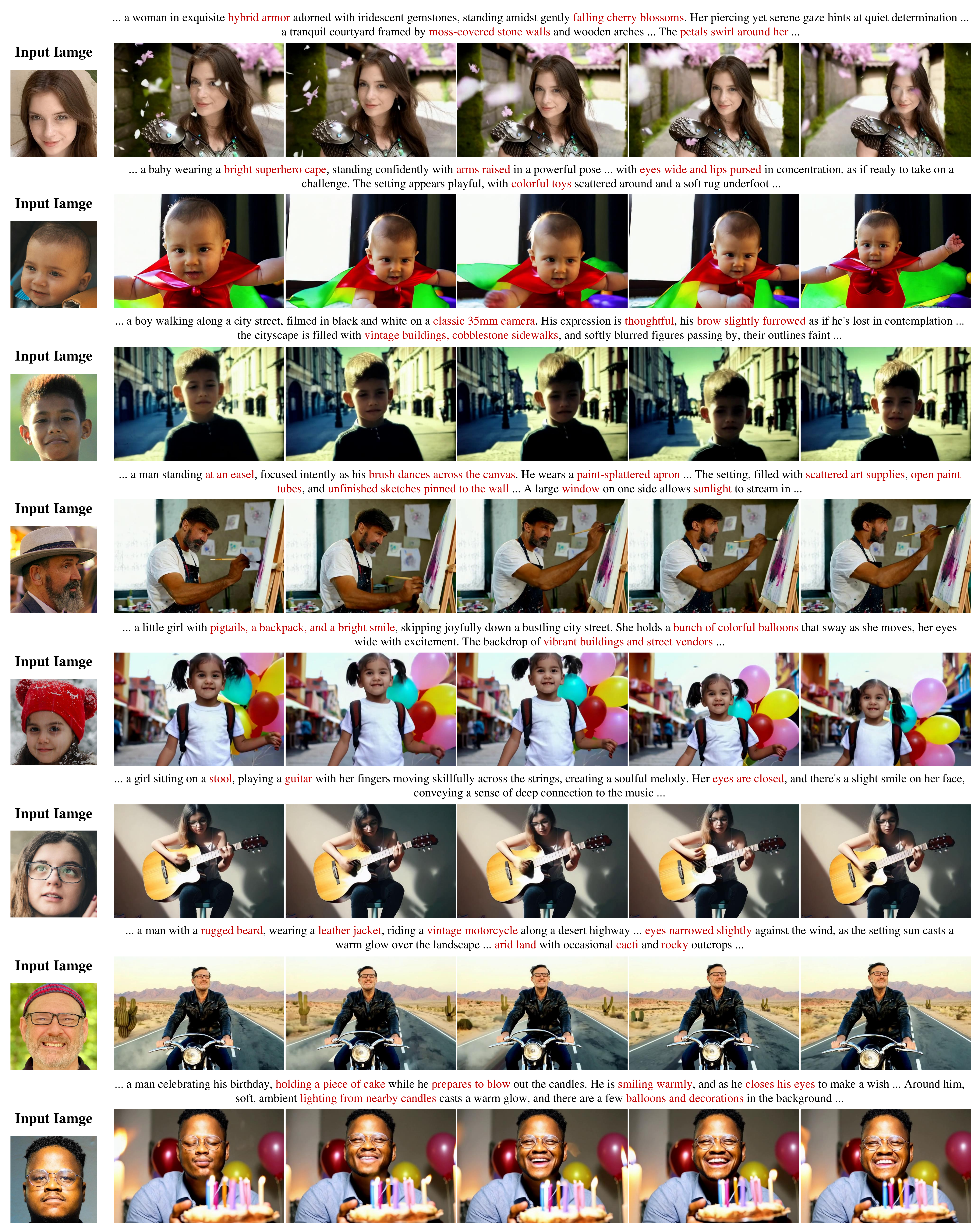}
    \caption{\textbf{More Showcases of Identity-Preserving Videos Generated by ConsisID.}}
    \label{fig: main results 4}
\end{figure*}

Due to space constraints, to assess the robustness and generalizability of our method, we present more ID-preserving video generation results in Figures \ref{fig: main results 2}, \ref{fig: main results 3} and \ref{fig: main results 4}, covering different people and different text prompts. ConsisID not only generates faces that match the identity of the reference image but also adheres to the text prompt, allowing for control over the character’s expressions, attire, actions, age, background, and even camera angles (\eg, detailed close-ups, wide panoramic views). These results substantiate the effectiveness of the Global / Local Facial Extractors, and the Consistency Training Recipe can enhance performance.

\section{Additional Experimental Details}
\subsection{Visualization of Different Injection Methods}\label{sec: Visualization of Different Injection Methods}
\begin{figure}[!t]
    \centering
    \includegraphics[width=1\linewidth]{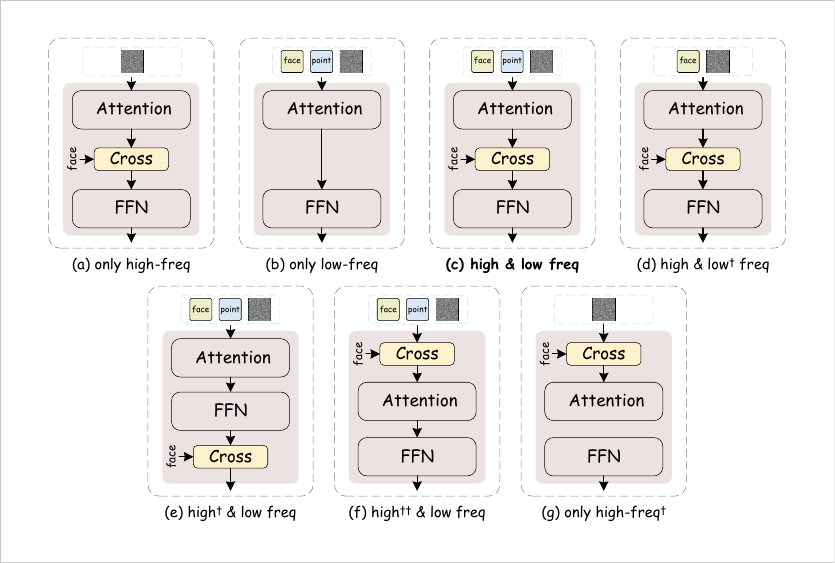}
    \caption{\textbf{Visualization of Different Methods of Injecting Control Signals into DiT.} Only (c), which injects high-frequency information between the Attention and FFN modules, while incorporating low-frequency signals, including facial key points, into the shallow layers of the network, resulting in optimal performance.}
    \label{fig: visualization_of_injection}
\end{figure}

To enhance the explanation of \textit{Identity Signal Injection in DiT} presented in the main text, we visualized various schemes, as shown in Figure \ref{fig: visualization_of_injection}. For simplicity, the visualization of the text branch is omitted. ConsisID employs scheme (c), which injects high-frequency information between the Attention and FFN modules, while integrating low-frequency signals (with facial key points) into the shallow layers of the network, achieving optimal result.

\subsection{Validation of the Automatic Metrics}\label{sec: Validation of the Automatic Metrics}
\begin{figure}[ht]
    \centering
    \includegraphics[width=0.9\linewidth]{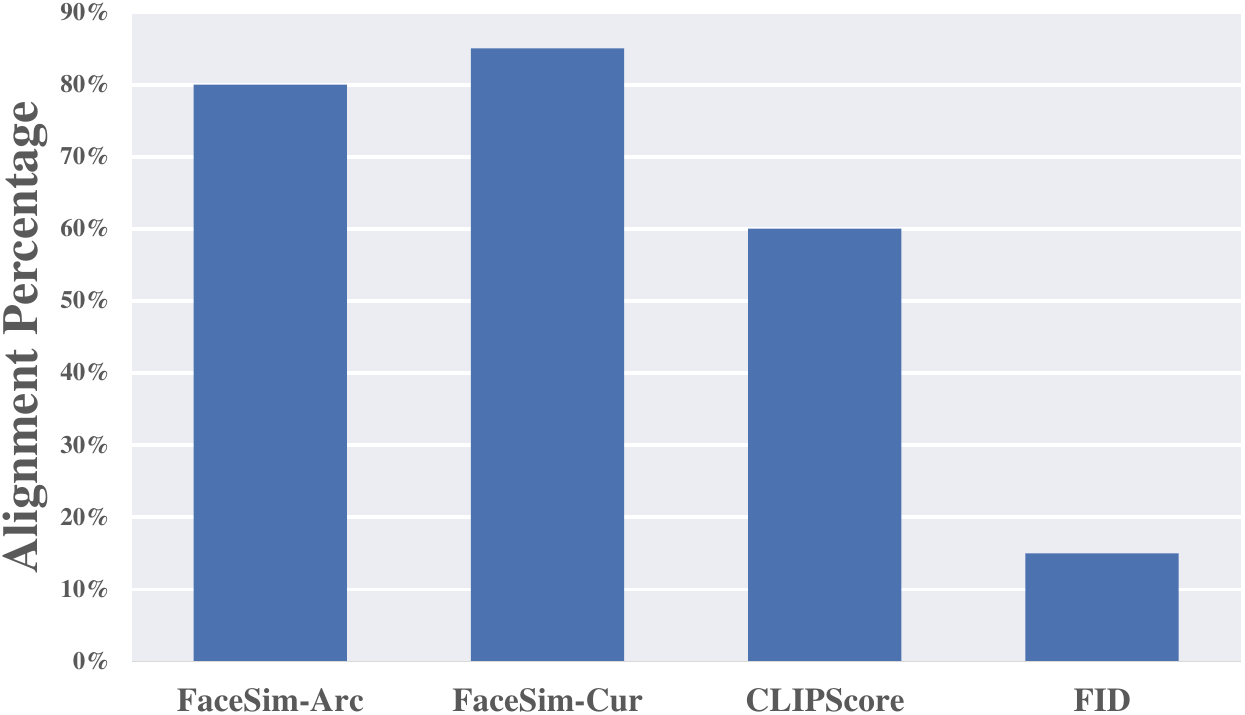}
    \caption{\textbf{Cross Validation between Automatic Metrics and Human Perception.} Existing metrics show limited alignment with human, particularly CLIPScore and FID.}
    \label{fig: metrics_alignment}
\end{figure}

In order to assess the effectiveness of the different metrics, we preform a cross-validation using the results of the user study. Specifically, we obtain FaceSim-Arc \cite{arcface}, FaceSim-Cur, CLIPScore \cite{clipscore} and FID \cite{FID} scores for each video in the questionnaire. We then identify the highest scoring option for each metric and compared these results with the questionnaire responses, as shown in Figure \ref{fig: metrics_alignment}. Although CLIPScore \cite{clipscore} reflect model performance reasonably well, their alignment with human perception remains limited. In particular, FID \cite{FID} showed an inverse relationship with human perception, with the lower quality ID-Animator \cite{ID-Animator} receiving a higher score. Therefore, the quantitative results presented in the main text should be interpreted cautiously.

\subsection{Details of Resource}\label{sec: Details of Resource}
We employ Nvidia H100 (x40) and A100 (x8) for all the experiments. All implementations are conducted on the basis of the official code using the PyTorch framework.

\subsection{Details of Evaluation Models}\label{sec: Details of Evaluation Models}
\myparagraph{Vidu \cite{Vidu}.}
Vidu1.5 is currently the only closed-source model supporting tuning-free IPT2V. It can generate videos of 4 or 8 seconds in length, with resolutions of 480p, 720p, or 1280p, and aspect ratios of 16:9, 9:16, or 1:1. We used its official default settings to generate 4-second, 480p, 16:9 videos for best comparison.

\noindent\myparagraph{ID-Animator \cite{ID-Animator}.}
ID-Animator is the sole open-source model currently supporting tuning-free IPT2V. It utilizes a UNet-based architecture and is designed to generate 2-second (16-frame) videos at a resolution of 512$\times$512. We utilized its official default configuration of DreamBooth (realisticVisionV60B1) to generate videos for comparison.

\noindent\myparagraph{DreamVideo \cite{dreamvideo}.}
DreamVideo is an open-source model supporting tuning-based IPT2V. It utilizes a UNet-based architecture and is designed to generate 4-second (32-frame) videos at a resolution of $256 \times 256$. For fairness, we use only one reference image and the official default settings (e.g., steps, learning rate) to train the model, and then generate videos for comparison.

\noindent\myparagraph{MotionBooth \cite{motionbooth}.}
MotionBooth is an open-source model supporting tuning-based IPT2V.  It utilizes a UNet-based architecture and is designed to generate 576$\times$320$\times$24 and 512$\times$320$\times$16 videos, respectively. For fairness, we use only one reference image and the official default settings (e.g., 512$\times$320$\times$16, steps) to train the model, and then generate videos for comparison. Due to the requirement to specify the direction of camera movement, we fix the camera to move to the left.

\noindent\myparagraph{Magic-Me \cite{magic-me}.}
Magic-Me is an open-source model supporting tuning-based IPT2V. It utilizes a UNet-based architecture and is designed to generate 8-second (16-frame) videos at a resolution of 1024$\times$1024. For fairness, we use only one reference image and the official default settings (e.g., steps, learning rate) to train the model, and then generate videos for comparison.

\noindent\myparagraph{CogVideoX \cite{cogvideox}, EasyAnimate \cite{easyanimate}, OpenSora-Plan \cite{opensoraplan}, DynamiCrafter \cite{Dynamicrafter}.} These models are open-source foundational generation models supporting image-to-video generation. While CogVideoX, EasyAnimate, and OpenSora-Plan are based on DiT architecture, DynamiCrafter employs a UNet-based architecture. Due to the lack of support for IPT2V in all these models, the process begins by generating the initial frame using PhotomakerV2 \cite{photomaker}. Subsequently, the respective models are used to generate the subsequent frames: CogVideoX-5B-I2V, EasyAnimate v4, OpenSora-Plan v1.3, and DynamiCrafter512. Due to the differences in supported resolution and length, we use the official default settings to ensure optimal performance.

\subsection{Details of Implementation}\label{sec: Additional Implementation Details}
\textbf{(1)} The section of \textit{Quantitative Analysis} requires each model to generate 13,500 videos (30$\times$5$\times$90). To minimize computational overhead, we select only 2 reference images per ID, each with 90 text prompts in the section \textit{Effect of the Identity Signal Injection in DiT} (30$\times$2$\times$90); 60 IDs each with 1 text prompt in the section \textit{Comparison with Tuning-based Methods} and \textit{Ablation on the Number of Inference Steps} (60$\times$1$\times$1); and select only 1 reference image per ID, each with 45 text prompts for the remaining experiments (30$\times$1$\times$45), including the \textit{Comparison with I2V Methods}, \textit{Fine-grained Ablation Study}, etc. \textbf{(2)} For all the baselines used in this paper, including Vidu \cite{Vidu}, ID-Animator \cite{ID-Animator}, DreamVideo \cite{dreamvideo}, MotionBooth \cite{motionbooth} and Magic-Me \cite{magic-me}, we use the default settings from their official websites (\eg, resolution, fps, inference steps, training steps, etc.) to ensure optimal results. For MotionBooth \cite{motionbooth}, since it requires the direction of camera movement to be specified, we set the camera to move to the left, which is also the official setting. \textbf{(3)} The evaluation dataset consists of 30 individuals, including celebrities, ordinary people, and those of diverse skin tones, as demonstrated by the qualitative results presented in this paper. This diversity enhances the comprehensiveness and reliability of the experimental data. Furthermore, the text prompts cover a wide range of expressions, actions, and backgrounds, providing a thorough assessment of the generalizability of IPT2V.

\begin{figure}[!t]
    \centering
    \includegraphics[width=0.9\linewidth]{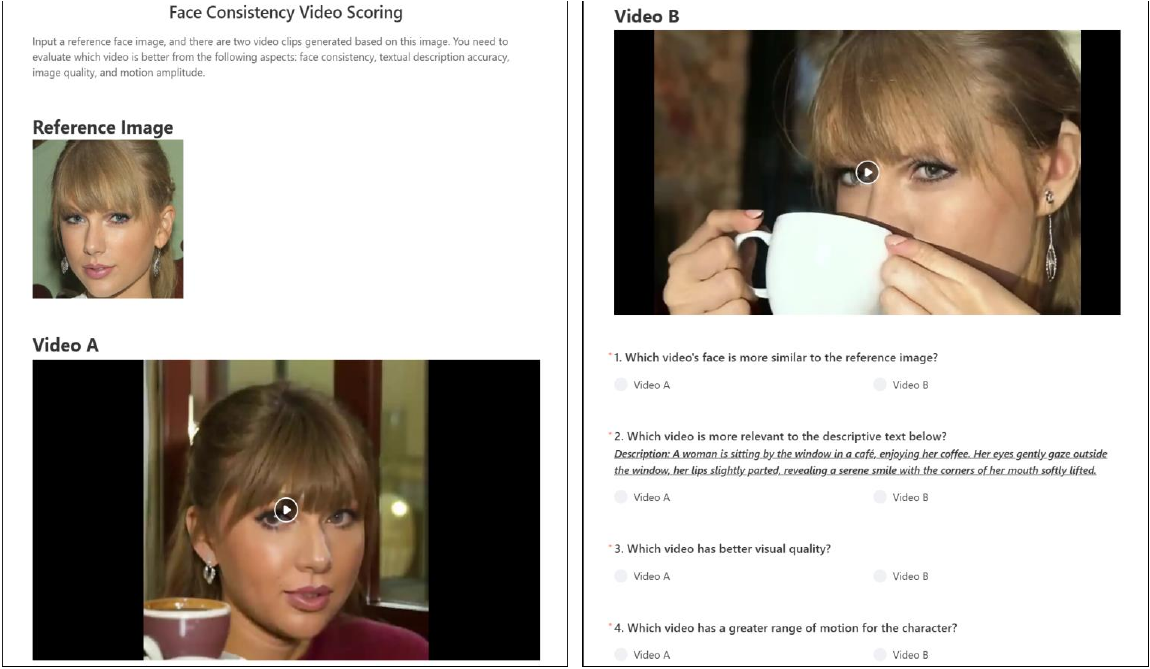}
    \caption{\textbf{Visualization of the Questionnaire for User Study.}}
    \label{fig: questionnaire_visualization}
\end{figure}

\subsection{Details of Human Evaluation}\label{sec: additional details of human evaluation}
To illustrate the user study intuitively, we provide a visualisation of the questionnaire in Figure \ref{fig: questionnaire_visualization}. In addition, to increase the reliability and diversity of the questionnaire, we implement the following rules:
\begin{itemize}
    \item The presentation order of different videos is randomized to reduce cognitive bias among respondents.
    \item A sliding verification upon submission is required to confirm that all responses are submitted manually and not by automated bots.
    \item Each IP address is restricted to a single submission, and users are required to log in before voting to ensure each individual could submit only once.
    \item Questionnaires where 90\% of responses selected the same option (all A or all B) are discarded.
    \item The primary voting population consisted of undergraduate, master’s, and PhD students from universities, along with a portion of the general public outside the field.
    \item The validity of data is assessed based on the time spent completing the questionnaire; responses with completion times of less than two minutes are excluded, as it typically takes 2--5 minutes to complete.
    \item Validity is also evaluated based on response distribution. Since the A/B options are randomly assigned for each question, we discarded questionnaires where 90\% of responses favored a single option (either all A or all B).
\end{itemize}

\section{Additional Statement}\label{sec: Additional Statement}
\subsection{Support for Key Findings}  These findings are not merely our observations but are synthesized from and validated by existing diffusion and ViT literature, with additional support provided by our experiments in Main Sec. \ref{sec: Effect of the Identity Signal Injection in DiT}. \textbf{Finding 1)} [\cite{rethinking_DiT} (Sec. 3.2), \cite{Freeu} (Sec. 1)] highlight that diffusion models’ noise prediction is fundamentally low-level and benefits from a U-Net bias. In Main Sec. \ref{sec: Effect of the Identity Signal Injection in DiT}, Main Table \ref{tab: validation_on_id_signal_injection} (f–g) shows training instability when low-frequency features are omitted, reinforcing their crucial role. \textbf{Finding 2)} [\cite{yu2024promptfix} (Sec. 1)] shows the importance of high-frequency features in the diffusion model. [\cite{U-ViT} (Abs), \cite{ViT} (Sec. 1)] note that vision transformers’ challenges with capturing high-frequency detail. 
Convergent evidence in Main Figure \ref{fig: injection_of_frequency} (f) indicates transformers’ high-frequency handling warrants deeper investigation. In Main Sec. \ref{sec: Effect of the Identity Signal Injection in DiT}, Main Figure \ref{fig: injection_of_frequency} confirms improvements when decoupling high- and low-frequency signals, despite a legend error we have corrected. 

\subsection{Justification of the Fine-grained Feature}
For how to ensure that the feature output by Local Facial Extractor remains fine-grained: Q-Former serve as a fusion mechanism rather than an extractor. Among the input into it, Face Extractor, as a face recognition backbone, plays a central role by inherently extracting fine-grained features invariant to non-identity attributes (e.g., expression, posture). CLIP provides secondary semantic features for editing, while Dropout \cite{drop_token, dropout} are employed to it to maintain the Q-Former's fine-grained feature dominance. Main Table \ref{tab: validation_on_id_signal_injection} shows both modules are distinct and complementary.

\subsection{The feature of Diffusion Transformer}
Sora \cite{SORA}, based on the DiT architecture, exhibits significant potential in simulating the physical world. Recently, foundational models for visual generation have shifted from UNet \cite{animatediff, Dynamicrafter} to DiT \cite{allegro, opensora, movie_gen, cogvideox, easyanimate}, owing to the latter's scalability and superior performance. Accordingly, our ConsisID is based on DiT instead of UNet architecture.

\subsection{Generalization to Image Generation}
Despite being trained exclusively on video data, ConsisID can leverage the generalization capabilities of CogVideoX \cite{cogvideox} to generate high-quality, identity-preserving images. This is achieved by either setting the $frame$ parameter to 1 or extracting the first frame of a video as an image.

\subsection{Ethics Statement}
ConsisID is capable of generating high-quality, realistic human videos. However, it also presents potential negative impacts, as the technique may be utilized to produce deceptive content for fraudulent activities. It is important to recognize that any technology is susceptible to misuse \cite{yan2023ucf, yan2024df40, yan2024generalizing, yan2024transcending}.

\subsection{Reproducibility Statement}
First, we have explained the implementation of ConsisID in detail in section \ref{sec: method}. Second, we have explained the details of training and inference in section \ref{sec: experiment}. Finally, the data and codes used in this work will be open-source online.

\subsection{Copyright Statement} 
The training data is sourced from in-house datasets, and only a subset of the data (CC BY 4.0 license) will be made publicly available. The video content exclusively features humans, and any NSFW content is detected and removed based on the video captions. The videos come from different regions of the world to ensure they are representative.

\subsection{Limitations and Future Work} Existing metrics fail to accurately assess the capabilities of different ID preservation models. While ConsisID can generate realistic and natural videos based on a text prompt, commonly used metrics such as CLIPScore \cite{clipscore} and FID \cite{FID} exhibit minimal differences compared to previous methods. A promising approach is to develop a metric that better aligns with human perception.

% \subsection{Limitations and Future Work}\label{sec: limitation and future work}
% \textbf{(1).} Existing metrics do not accurately measure the capabilities of different ID preservation models. Although ConsisID can generate realistic and natural videos following a text prompt, metrics such as CLIPScore \cite{clipscore} and FID \cite{FID} show little difference from previous methods. A viable direction is to find a metric that more closely resembles human perception. \textbf{(2).} We can only open source a portion of our in-house datasets.

\end{document}